\documentclass[10pt,twocolumn,letterpaper]{article}

\usepackage{cvpr}
\usepackage{times}
\usepackage{epsfig}
\usepackage{graphicx}
\usepackage{amsmath}
\usepackage{amssymb}
\usepackage{graphicx}
\usepackage{subfigure}
\usepackage{color}
\usepackage{tabularx} % tabularx
\usepackage{booktabs} % \toprule, \midrule, \bottomrule
\usepackage{array}
\usepackage{multirow}
\usepackage[table]{xcolor}
\usepackage{docmute}
\usepackage{pdfpages}
\usepackage{fancyhdr}

\DeclareMathOperator*{\diag}{diag}
\DeclareMathOperator*{\determinat}{det}
\DeclareMathOperator*{\trace}{trace}
\DeclareMathOperator*{\arctantwo}{arctan2}

\usepackage[breaklinks=true,bookmarks=false,colorlinks]{hyperref}

\cvprfinalcopy % *** Uncomment this line for the final submission

\def\cvprPaperID{2778} % *** Enter the CVPR Paper ID here

\begin{document}
	
	%%%%%%%%% TITLE
	\title{Minimal Solutions for Relative Pose with a Single Affine Correspondence}
	
	\author{Banglei Guan$^{1}$,\ \ Ji Zhao\thanks{Corresponding author.} ,\ \ Zhang Li$^{1}$,\ \ Fang Sun$^{1}$ and \ \ Friedrich Fraundorfer$^{2,3}$\\
		 {\small$^{1}$College of Aerospace Science and Engineering, National University of Defense Technology, China} \\{\small$^{2}$Institute for Computer Graphics and Vision, Graz University of Technology, Austria} \\{\small$^{3}$Remote Sensing Technology Institute, German Aerospace Center, Germany}\\
		{\tt\small guanbanglei12@nudt.edu.cn \quad zhaoji84@gmail.com \quad zhangli{\underline{ }}nudt@163.com}\\
		{\tt\small sunfang@nudt.edu.cn \quad fraundorfer@icg.tugraz.at}
	}
	
	\maketitle
%	\thispagestyle{empty}
	
	%%%%%%%%% ABSTRACT
	\begin{abstract}
		In this paper we present four cases of minimal solutions for two-view relative pose estimation by exploiting the affine transformation between feature points and we demonstrate efficient solvers for these cases. It is shown, that under the planar motion assumption or with knowledge of a vertical direction, a single affine correspondence is sufficient to recover the relative camera pose. The four cases considered are two-view planar relative motion for calibrated cameras as a closed-form and a least-squares solution, a closed-form solution for unknown focal length and the case of a known vertical direction. These algorithms can be used efficiently for outlier detection within a RANSAC loop and for initial motion estimation. All the methods are evaluated on both synthetic data and real-world datasets from the KITTI benchmark. The experimental results demonstrate that our methods outperform comparable state-of-the-art methods in accuracy with the benefit of a reduced number of needed RANSAC iterations. 
		
	\end{abstract}
	
	%%%%%%%%% BODY TEXT
	%-------------------------------------------------------------------------
	\section{Introduction}
	
	Simultaneous localization and mapping (SLAM), visual odometry (VO) and Structure-from-Motion (SfM) have been active research topics in computer vision for decades~\cite{scaramuzza2011visual, schoenberger2016sfm}. These technologies have been used successfully in a wide variety of applications and they play an important role in future technologies like autonomous driving. Relative pose estimation from two views is regarded as a fundamental algorithm, which is an essential part of SLAM and SfM pipelines. %In SfM pipelines two-view relative pose estimation is e.g. used in the initial process of creating the epipolar graph which determines the order in which the images are processed. 
	Thus, improving the accuracy, efficiency and robustness of relative pose estimation algorithms is still of relevant interest~\cite{Agarwal2017,barath2018five,Silveira_2019_CVPR,zhao2019efficient}. 
	
	\begin{figure}[ht]
		\begin{center}
			\includegraphics[width=1.0\linewidth]{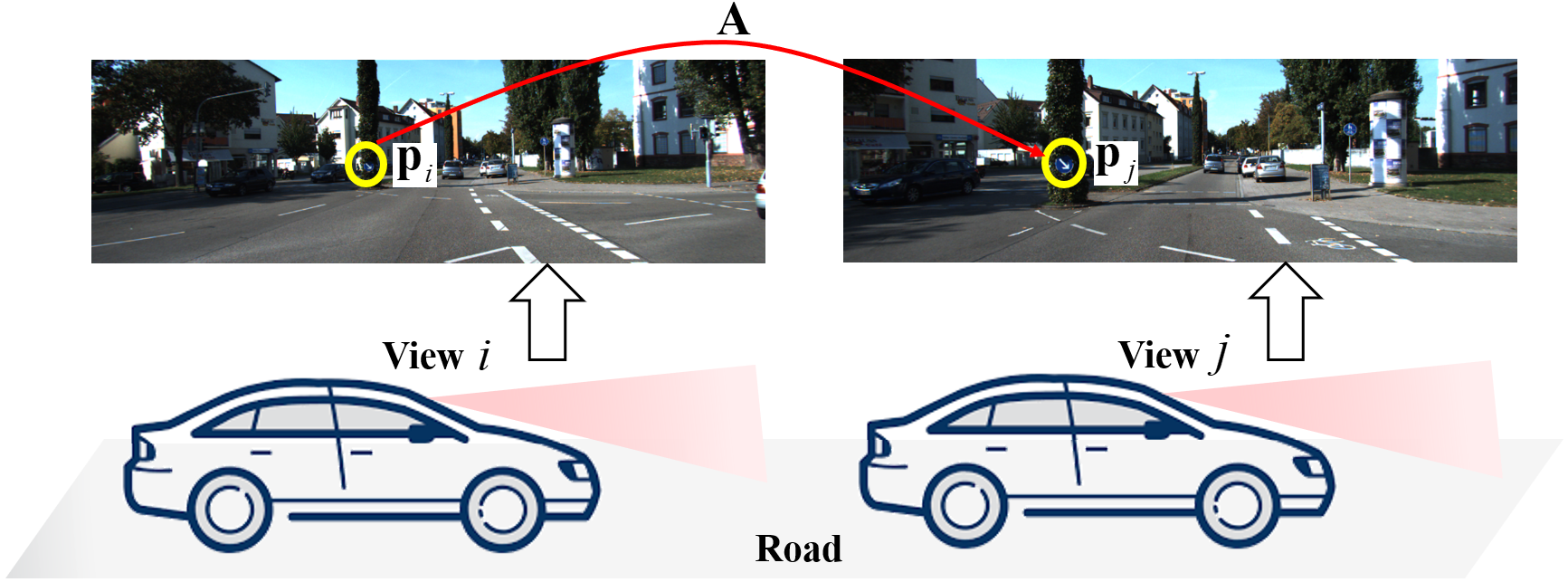}
		\end{center}
		\caption{An affine correspondence between two cameras. The local affine transformation $\mathbf{A}$ transforms the patches surrounding of point correspondence ($\mathbf{p}_{i}$, $\mathbf{p}_{j}$).}
		\label{fig:AffineTransformation}
		\vspace{-15pt}
	\end{figure}   
	
	Most of the SLAM and SfM pipelines follow the scheme where 2D-2D putative correspondences between subsequent views are established by feature matching. Then a robust motion estimation framework such as the Random Sample Consensus (RANSAC)~\cite{fischler1981random} is typically adopted to identify and remove matching outliers. Finally, only inlier matches between subsequent views are used to estimate the final relative pose~\cite{scaramuzza2011visual}. This outlier removal step is critical for the robustness and reliability of the pose estimation step. Besides, the efficiency of the outlier removal process affects the real-time performance of SLAM and SfM directly, in particular, as the computational complexity of the RANSAC estimator increases exponentially with respect to the number of data points needed. Thus minimal case solutions for relative pose estimation are still of significant importance~\cite{barath2017minimal,barath2018five,zhao2019minimal,Duff_2019_ICCV}.
	
	% for non-calibrated images, for calibrated images, for plane scenario and calibrated images
	%If there exists a dominant plane present in the scene, the four-point homography estimation for calibrated images is also widely used~\cite{HartleyZisserman-472}. 
	The idea of minimal solutions for relative pose estimation ranges back to the work of Hartley and Zisserman with the seven-point method~\cite{HartleyZisserman-472}. Other classical works are the five-point method ~\cite{nister2004efficient} and the homography estimation method~\cite{HartleyZisserman-472}. 
	By exploiting motion constraints on camera movements or utilizing an additional sensor like an inertial measurement unit (IMU), the minimal number of point correspondences needed can be further reduced, which makes the outlier removal more efficient and numerically more stable. For instance, two points are sufficient to recover camera motion under the planar motion assumption since the pose only has two degrees of freedom (DOF)~\cite{ortin2001indoor,choi2018fast,chou2018two}. Another example is to make use of the Ackermann steering principle which allows us to parameterize the camera motion with only one point correspondence~\cite{scaramuzza2009real,huang2019motion}. These scenarios are typical for self-driving vehicles and ground robots. For unmanned aerial vehicles (UAV) and smartphones, a camera is often used in combination with an IMU~\cite{Guan2018Minimal}. The partial IMU measurements can be used to provide a known gravity direction for the camera images. In this case relative pose estimation is thus possible with only three point correspondences.~\cite{fraundorfer2010minimal,naroditsky2012two,sweeney2014solving,SaurerVasseur-457}. 
	
	% FF, I removed this as the paper does not talk about points + scale or rotation anymore.
	%The aforementioned methods only use the point coordinates for relative pose estimation. However, the widely-used feature detectors, such as SIFT~\cite{Lowe2004Distinctive}, SURF~\cite{Bay2008346} and ORB~\cite{Rublee11}, not only provide the point coordinates, but also a rotation and a scale. This additional information can be also exploited for relative pose estimation, which reduces the number of required points as well~\cite{liwicki17scale,barath2018five}. 
	
	It is now possible to replace simple point correspondences with  affine-covariant feature detectors, such as ASIFT~\cite{morel2009asift} and MODS~\cite{mishkin2015mods}. Such an affine correspondence (AC) consists of a point correspondence and a $2\times2$ affine transformation, see Figure~\ref{fig:AffineTransformation}. It has been proven that 1~AC yields three constraints on the geometric model estimation~\cite{bentolila2014conic, raposo2016theory, barath2018efficient}. In this paper we exploit these additional affine parameters in the process of relative pose estimation which allows to reduce the number of correspondences needed. We propose the following $4$ novel minimal solutions for relative pose estimation using a single affine correspondence:
	\vspace{-5pt}
	\begin{itemize}
		\item Three solvers under the planar motion constraint are proposed. We prove that a single affine correspondence is sufficient to recover the planar motion of a calibrated camera (2DOF) and a partially uncalibrated camera for which only the focal length is unknown (3DOF).
		\vspace{-5pt}  
		\item A fourth solver for the case of a known vertical direction is proposed. The egomotion estimation of calibrated camera with a common direction has 3DOF, and we will show that only a single affine correspondence is required to estimate the relative pose for this case.
	\end{itemize}
	\vspace{-5pt}

	The remainder of the paper is organized as follows. First we review related work in Section~\ref{sec:relatedwork}. We propose three minimal solutions for planar motion estimation in Section~\ref{sec:planarmotion}. In Section~\ref{sec:knownverticaldirection}, we propose a minimal solution for two-view relative motion estimation with known vertical direction. In Section~\ref{sec:experiments}, we evaluate the performance of proposed methods using both synthetic and real-world dataset. Finally, concluding remarks are given in Section~\ref{sec:conclusion}.
	
	%-------------------------------------------------------------------------
	\section{\label{sec:relatedwork}Related Work}
	
	For non-calibrated cameras, a minimum of 7 point correspondences is sufficient to estimate the fundamental matrix~\cite{HartleyZisserman-472}. If the camera is partially uncalibrated such that only the common focal length is unknown, a minimum of 6 point correspondences is required to estimate the relative pose~\cite{stewenius2005minimal,kukelova2017clever}. For calibrated cameras, at least 5 point correspondences are needed to estimate the essential matrix~\cite{nister2004efficient}. If all the 3D points lie on a plane, the point correspondences are related by a planar homography and the number of required point correspondences is reduced to 4~\cite{HartleyZisserman-472}. The relative pose of two views can be recovered by the decomposition of the essential matrix or the homography.   
	
	To further improve the computational efficiency and reliability of relative pose estimation, assumptions about the camera motion or additional information can help to reduce the number of required point correspondences across views. For example, if the camera is mounted on ground robots and follows planar motion, the relative pose of two views has only 2DOF and can be estimated by using 2 point correspondences~\cite{ortin2001indoor,choi2018fast,chou2018two}. By taking into account the Ackermann motion model, only 1 point correspondence is sufficient to recover the camera motion~\cite{scaramuzza2009real}. 
	
	%By contrast, 
	%, because the rotation of camera can be obtained from the IMU information.
	When additional information can be provided by an additional sensor, such as an IMU, the DOF of relative pose estimation can also be reduced. If the rotation of the camera is fully provided by an IMU, only the translation of two views is unknown and can easily be solved with 2 point correspondences~\cite{kneip2011robust}. 
	It is more often the case that a common direction of rotation is assumed to be known. This common direction can be determined from an IMU (which provides the known pitch and roll angles of the camera), but as well from vanishing points extracted across the two views. 
	When the common direction of rotation is known, a variety of algorithms have been proposed to estimate the relative pose utilizing this information~\cite{fraundorfer2010minimal,naroditsky2012two,sweeney2014solving, SaurerVasseur-457,guan2018visual, ding2019efficient}. %For the common 3D scene, a simplified essential matrix is derived for two views with a common direction and solved with 3 point correspondences~\cite{fraundorfer2010minimal}. To further reduce the number of point correspondences, several simplified homographies are also derived for the 2D scene and solved by using a minimum of 1.5 point correspondences~\cite{guan2018visual} and 2 point correspondences~\cite{SaurerVasseur-457}.
	
	Recently, a number of methods have been proposed which reduce the number of required points by exploiting the additional affine parameters between two feature matches. These additional information can come from the feature's rotation and scale estimates when SIFT~\cite{Lowe2004Distinctive} or SURF~\cite{Bay2008346} feature detectors are used.  From five such point correspondences extended by the rotational angles of the features the fundamental matrix can be computed~\cite{barath2018five}. Similarly, the homography can be estimated by using two correspondences when including the corresponding rotational angles and scales of the features~\cite{barath2019homography}. Of high interest are methods which use affine correspondences obtained by an affine-covariant feature detector, such as ASIFT~\cite{morel2009asift} and MODS~\cite{mishkin2015mods}.  One AC yields three constraints on the geometric model estimation. This allows the estimation of a fundamental matrix from 3 ACs~\cite{bentolila2014conic}. The estimation of a homography and an essential matrix can be accomplished from 2 ACs~\cite{raposo2016theory,eichhardt2018affine,barath2018efficient}. There is an independent work which also uses a single AC to estimate relative planar motion~\cite{hajder2019relative}. Furthermore, it is shown in~\cite{raposo2016theory} that ACs have benefits as compared to point correspondences for visual odometry in the presence of many outliers. 
	
	%-------------------------------------------------------------------------
	\section{\label{sec:planarmotion}Relative Pose Estimation Under Planar Motion}
	
\begin{figure}[ht]
		\vspace{-6pt}
		\begin{center}
			\includegraphics[width=0.55\linewidth]{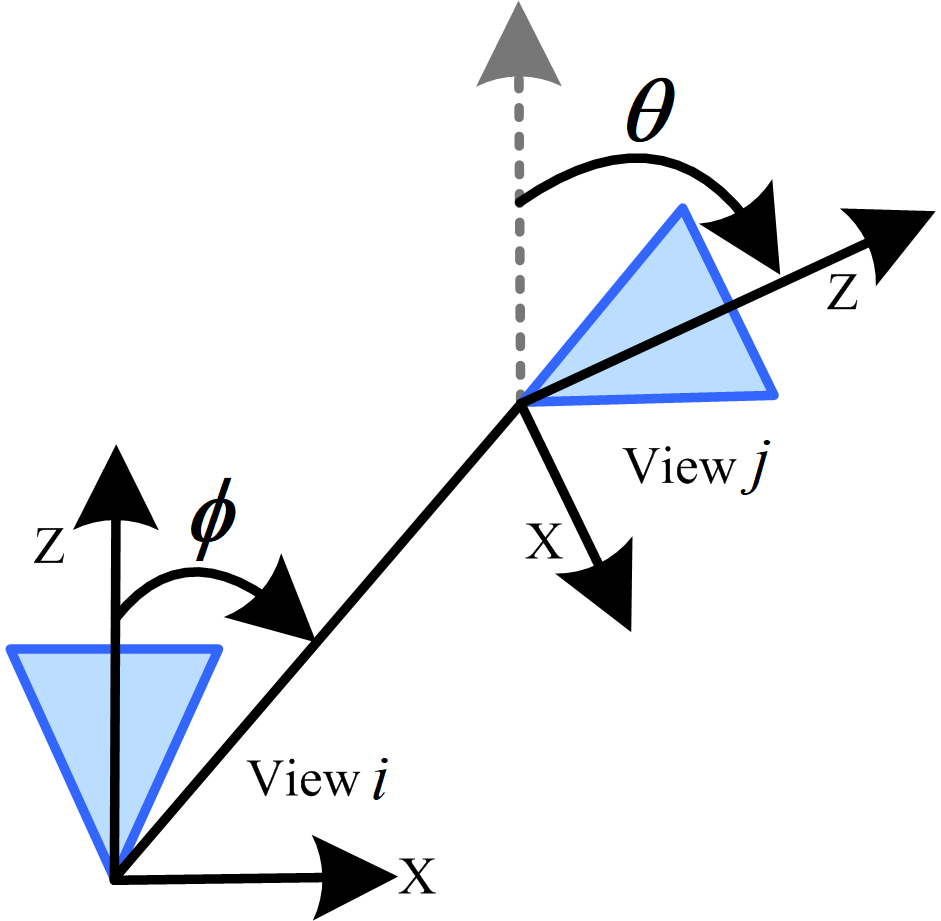}
		\end{center}
		\vspace{-10pt}	
		\caption{Planar motion between two cameras in top-view. There are two unknowns: yaw angle $\theta$ and translation direction $\phi$.}
		\label{fig:PlanarMotion}
		\vspace{-5pt}	
	\end{figure}

	For planar motion shown in Figure~\ref{fig:PlanarMotion}, we derive three minimal solvers by exploiting one affine correspondence only. (1) We develop two minimal solvers for calibrated cameras. Since one AC provides three independent equations and there are two unknowns for the pose, the equation system is over-determined. We propose two variants for this scenario including a closed-form solution and a least-squares solution. (2) For uncalibrated cameras with unknown focal length only, we propose a minimal solver for this scenario as well.
	
	%-------------------------------------------------------------------------
	\subsection{\label{sec:1ACplane}Solver for Planar Motion with Calibrated Camera}
	%\vspace{-5pt}
	
	With known intrinsic camera parameters, the epipolar constraint between views $i$ to $j$ is given as follows~\cite{HartleyZisserman-472}:
	%\vspace{-5pt}
	\begin{equation}
	\mathbf{p}_{j}^T\mathbf{E}\mathbf{p}_{i} =0, 
	\label{eq:epipolar geometry}
	%\vspace{-2pt}
	\end{equation}
	where $\mathbf{p}_{i}=[u_{i},v_{i},1]^{T}$ and $\mathbf{p}_{j}=[u_{j},v_{j},1]^{T}$ are the normalized homogeneous image coordinates of a feature point in views $i$ and $j$, respectively. $\mathbf{E}=[\mathbf{t}]_{\times}\mathbf{R}$ is the essential matrix, where $\mathbf{R}$ and $\mathbf{t}$ represent relative rotation and translation respectively.
	
	For planar motion, we assume that the image plane of the camera is vertical to the ground plane without loss of generality, see Figure~\ref{fig:PlanarMotion}. There are only a Y-axis rotation and 2D translation between two different views, so the rotation $\mathbf{R}=\mathbf{R}_{y}$ and the translation $\mathbf{t}$ from views $i$ to $j$ can be written as: 
	{\small\begin{equation}
		\mathbf{R}_{y} = \begin{bmatrix}{\cos{(\theta)}}&0&{-\sin{(\theta)}}\\
		0&1&0\\
		{ \sin{(\theta)}}&0&{\cos{(\theta)}}
		\end{bmatrix}, \ \	\mathbf{t} = -\mathbf{R}_{y}\begin{bmatrix}
		{\rho\sin{(\phi)}}\\
		{0}\\
		{\rho\cos{(\phi)}}
		\end{bmatrix}.
		\label{eq:Ry1T1}
		\end{equation}}\\
	where $\rho$ is the distance between views $i$ and $j$. Based on Eq.~\eqref{eq:Ry1T1}, the essential matrix $\mathbf{E}=[\mathbf{t}]_{\times}\mathbf{R}_{y}$ under planar motion is reformulated:
	\begin{equation}
	\mathbf{E} = {\rho}\begin{bmatrix}
	{0}&{\cos{(\theta-\phi)}}&{0}\\
	{-\cos{(\phi)}}&{0}&{\sin{(\phi)}}\\
	{0}&{\sin{(\theta-\phi)}}&{0}
	\end{bmatrix}.
	\label{eq:essentialmatrix}
	\end{equation}
	
	By substituting the above equation into Eq.~\eqref{eq:epipolar geometry}, the epipolar constraint can be written as:
	\begin{equation}
	{v_i}{\sin(\theta-\phi)}+{v_i}{u_j}{\cos(\theta-\phi)} + {v_j}{\sin(\phi)} - {u_i}{v_j}{\cos(\phi)}=0. \label{eq:essentialequation1}
	\end{equation}
	
	Moreover, the widely-used affine-covariant feature detectors, \emph{e.g.} ASIFT~\cite{morel2009asift}, provide affine correspondences between two views directly. Here, we exploit the affine transformation in the relative pose estimation under planar motion, to further reduce the number of required point correspondences. Firstly, we introduce the affine correspondence, which is considered as a triplet: $(\mathbf{p}_{i},\mathbf{p}_{j},\mathbf{A})$. The local affine transformation $\mathbf{A}$ which relates the patches surrounding $\mathbf{p}_{i}$ and $\mathbf{p}_{j}$ is defined as follows~\cite{barath2018five}:
	\vspace{-6pt}
	\begin{equation}
	\begin{aligned}
	\mathbf{A} &=\begin{bmatrix}
	{{a_{11}}}&{{a_{12}}}\\
	{{a_{21}}}&{{a_{22}}}\\
	\end{bmatrix}.
	\end{aligned}
	\label{eq:affine transformation}
	\vspace{-6pt}
	\end{equation}

	The relationship of essential matrix $\mathbf{E}$ and local affine transformation $\mathbf{A}$ can be described as follows~\cite{barath2018efficient}
	\vspace{-4pt}
	\begin{equation}
	(\mathbf{E}^{T}\mathbf{p}_{j})_{(1:2)} = -(\mathbf{\hat{A}}^{T}\mathbf{E}\mathbf{p}_{i})_{(1:2)},
	\label{eq:E_Ac1}
	\vspace{-4pt}
	\end{equation}
	where $\mathbf{n}_i\triangleq{\mathbf{E}^{T}\mathbf{p}_{j}}$ and $\mathbf{n}_j\triangleq{\mathbf{E}\mathbf{p}_{i}}$ are the epipolar lines in the views $i$ and $j$, respectively. $\mathbf{\hat{A}}$ is a $3\times3$ matrix:
	\vspace{-4pt}
	\begin{equation}
	\mathbf{\hat{A}} = \begin{bmatrix}
	\mathbf{A}&\mathbf{0}\\
	\mathbf{0}&0
	\end{bmatrix}.
	\label{eq:Ac}
	\vspace{-4pt}
	\end{equation}

	By substituting Eq.~\eqref{eq:essentialmatrix} into Eq.~\eqref{eq:E_Ac1}, two equations which relate the affine transformation to the relative pose are obtained
	\vspace{-6pt}
	{\small\begin{align}
		&{a_{11}}{v_i}{\cos(\theta-\phi)} + {a_{21}}{\sin(\phi)} - ({a_{21}}{u_i}+{v_j}){\cos(\phi)}=0, \ \label{eq:essentialequation2}\\
		&{\sin(\theta-\phi)}+ ({a_{12}}{v_i}+{u_j}){\cos(\theta-\phi)}+ {a_{22}}{\sin(\phi)} -  \nonumber \\
		& \qquad\ \qquad\ \qquad\ \qquad\ \qquad\ \qquad\ \qquad \ \ \ \ {a_{22}}{u_i}{\cos(\phi)}=0. \label{eq:essentialequation3}
		\vspace{-4pt}
		\end{align}}
	
	\vspace{-28pt}
	\subsubsection{Closed-Form Solution}
	\vspace{-2pt}
	For an affine correspondence, the combination of Eqs.~\eqref{eq:essentialequation1}, \eqref{eq:essentialequation2} and \eqref{eq:essentialequation3} can be expressed as $\mathbf{C}\mathbf{x} = \mathbf{0}$, where $\mathbf{x} = [{\sin(\theta-\phi)},\; {\cos(\theta-\phi)},\; {\sin(\phi)},\; {\cos(\phi)}]^T$. 
	To facilitate the description of the following method, we denote 
	\vspace{-6pt}
	\begin{align}
	\begin{cases}
	x_1 \triangleq \sin({\theta-\phi}), \quad x_2 \triangleq \cos({\theta-\phi}) \\
	x_3 \triangleq \sin(\phi), \qquad \ \ \ x_4 \triangleq \cos(\phi) 
	\end{cases}
	\label{eq:triangeq}
	\end{align}
	
	\vspace{-6pt}
	By ignoring the implicit constraints between the entries of $\mathbf{x}$, \emph{i.e.}, $x_1^2 + x_2^2 = 1$ and $x_3^2 + x_4^2 = 1$, $\mathbf{x}$ should lie in the null space of $\mathbf{C}$. Thus the solution of the system $\mathbf{x}$ can be obtained directly based on the eigenvector of ${\mathbf{C}^T}\mathbf{C}$ corresponding to the least eigenvalue. Once $\mathbf{x}$ has been obtained, the angles $\phi$ and $\theta$ are 
	\vspace{-6pt}
	\begin{align}
	\begin{cases}
	{\phi}={\arctantwo({x_3},{x_4})},\\
	{\theta}={\arctantwo({x_1},{x_2})}+{\phi}.
	\end{cases}
	\label{eq:thetaphi}
	\vspace{-4pt}
	\end{align}
	\vspace{-24pt}
	\subsubsection{Least-Squares Solution}
	\vspace{-2pt}
	Eqs.~\eqref{eq:essentialequation1}, \eqref{eq:essentialequation2} and \eqref{eq:essentialequation3} together with the implicit constraints of the trigonometric functions can be reformulated as:
	\vspace{-8pt}
	\begin{align}
	\begin{cases}
	a_i x_1 + b_i x_2 + c_i x_3 + d_i x_4 = 0, \ i = 1, 2, 3 \\
	x_1^2 + x_2^2 = 1 \\
	x_3^2 + x_4^2 = 1
	\end{cases}
	\label{eq:reform1}
	\end{align}
	
	\vspace{-6pt}
	The coefficients $a_i$, $b_i$, $c_i$ and $d_i$ denote the problem coefficients in Eqs.~\eqref{eq:essentialequation1}, \eqref{eq:essentialequation2} and \eqref{eq:essentialequation3}. This equation system has $4$ unknowns and $5$ independent constraints, thus it is over-constrained. We find the least-squared solution by
	\vspace{-8pt}
	\begin{align}
	\label{equ:op_obj}
	\min_{\{x_i\}_{i=1}^4} & \ \ \sum_{i=1}^3 (a_i x_1 + b_i x_2 + c_i x_3 + d_i x_4)^2 \\
	\text{s.t.} & \ \ x_1^2 + x_2^2 = 1, \nonumber \ \ x_3^2 + x_4^2 = 1. \nonumber
	\vspace{-8pt}
	\end{align}
	
	The Lagrange multiplier method is used to find all stationary points in problem~\eqref{equ:op_obj}.
	The Lagrange multiplier is
	\vspace{-15pt}
	\begin{align}
	\label{equ:LagMul1}
	&L(x_1, x_2, x_3, x_4, \lambda_1, \lambda_2) \nonumber \\
	= & \sum_{i=1}^3 (a_i x_1 + b_i x_2 + c_i x_3 + d_i x_4)^2 \nonumber \\
	& + \lambda_1 (x_1^2 + x_2^2 - 1) + \lambda_2 (x_3^2 + x_4^2 - 1).
	\vspace{-8pt}
	\end{align}
	
	By taking the partial derivatives with $\{x_i\}_{i=1}^4$ and $\{\lambda_i\}_{i=1}^2$ and setting them to be zeros, we obtain an equation system with unknowns $\{x_i\}_{i=1}^4$ and $\{\lambda_i\}_{i=1}^2$, see the supplementary material. This equation system contains $6$ unknowns $\{x_1, x_2, x_3, x_4, \lambda_1, \lambda_2\}$, and the order is $2$. A Gr\"{o}bner basis solver with template size $42\times 50$ can be obtained by an automatic solver generator~\cite{larsson2017efficient}. It also shows that there are at most 8 solutions.
	
	%-------------------------------------------------------------------------
	\subsection{\label{sec:1ACUnkownF}Solver for Planar Motion and Unknown Focal Length}
	
	In this subsection, we assume that there is a camera with known intrinsic parameters except for an unknown focal length. This case is typical to be encounter in practice. For most cameras, it is often reasonable to assume that the cameras have square-shaped pixels and the principal point is well approximated by the image center~\cite{hartley2012efficient}. By assuming that the only unknown calibration parameter of the camera is the focal length $f$, the intrinsic matrix of the camera is simplified to $\mathbf{K} = \diag(f, f, 1)$.
	
	Since the intrinsic matrix is unknown, we can not obtain the coordinates of point features in the normalized image plane. Recall that the normalized homogeneous image coordinates of the points in views $i$ and $j$ are $\mathbf{p}_{i} = [u_i, v_i, 1]^T$ and $\mathbf{p}_{j} = [u_j, v_j, 1]^T$, respectively. Without loss of generality, we set the principle point as the centre of image plane. Denote  coordinates of a point in original image plane $i$ and $j$ as $\bar{\mathbf{p}}_i = [\bar{u}_i, \bar{v}_i, 1]^T$ and $\bar{\mathbf{p}}_j = [\bar{u}_j, \bar{v}_j, 1]^T$, respectively. We also denote $g = f^{-1}$ and obtain the following relations
	\vspace{-5pt}
	\begin{align}
	\begin{cases}
	u_i = f^{-1} \bar{u}_i = g \bar{u}_i, \qquad \	v_i = f^{-1} \bar{v}_i = g \bar{v}_i, \\
	u_j = f^{-1} \bar{u}_j = g \bar{u}_j, \qquad	v_j = f^{-1} \bar{v}_j = g \bar{v}_j. 
	\end{cases}
	\label{eq:coord}
	\end{align}
	
	By substituting Eq.~\eqref{eq:coord} into Eqs.~\eqref{eq:essentialequation1}, \eqref{eq:essentialequation2} and \eqref{eq:essentialequation3}, we also obtain three equations. To reduce the burden in notation, we substitute Eq.~\eqref{eq:triangeq} into the three equations. By combining them with two trigonometric constraints, we have a polynomial equation system as follows
	\vspace{-5pt}
	\begin{align}
	\begin{cases}
	{\bar{v}_i}g{x_1}+{\bar{v}_i}{\bar{u}_j}g^2{x_2} + {\bar{v}_j}g{x_3} - {\bar{u}_i}{\bar{v}_j}g^2{x_4}&=0 \\
	{a_1}{\bar{v}_i}g{x_2} + {a_3}{x_3} - ({a_3}{\bar{u}_i}+{\bar{v}_j})g{x_4}&=0 \\
	{x_1}+ ({a_2}{\bar{v}_i}+{\bar{u}_j})g{x_2} + {a_4}{x_3} - {a_4}{\bar{u}_i}g{x_4}&=0 \\
	x_1^2 + x_2^2 = 1 \\
	x_3^2 + x_4^2 = 1
	\end{cases}
	\label{eq:fundamentalequation2}
	\end{align}
	\vspace{-8pt}
	
	The above equation system contains $5$ unknowns $\{x_1, x_2, x_3, x_4, g\}$, and the order is $3$. The Gr\"{o}bner basis solver with template size $20\times 23$ can be obtained by an automatic solver generator~\cite{larsson2017efficient}. It also shows that there are at most $6$ solutions. Note that one trivial solution (${g}={x_1}={x_3}=0$, ${x_2}={x_4}=1$) can be safely removed considering $g= f^{-1}$ must be greater than 0.  
	
	%-------------------------------------------------------------------------
	\section{\label{sec:knownverticaldirection}Relative Pose Estimation with Known Vertical Direction}
	\vspace{-5pt}
	\begin{figure}[ht]
		\vspace{-10pt}
		\begin{center}
			\includegraphics[width=0.9\linewidth]{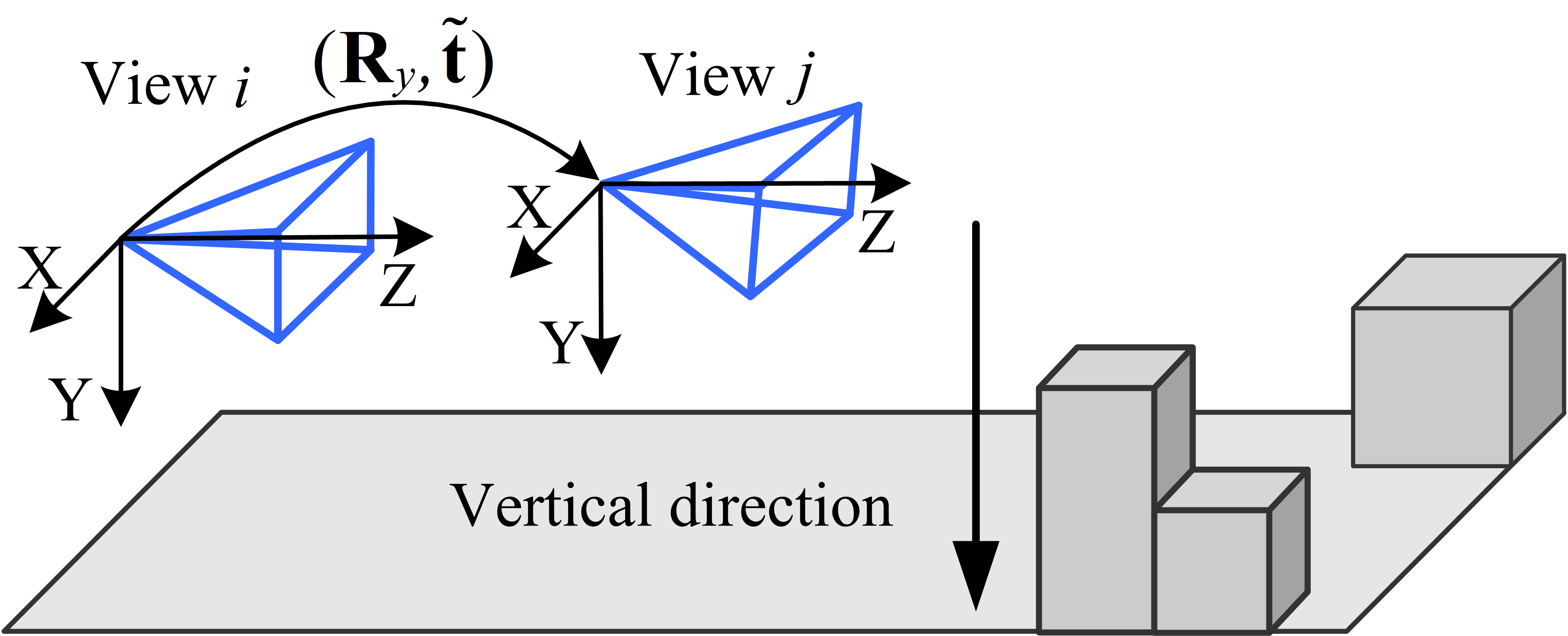}
		\end{center}
		\caption{Camera motion with known vertical direction. The unknowns include yaw angle $\theta$ and translation $[t_x,\; t_y,\; t_z]^T$.}
		\label{fig:knownverticaldirection}
		\vspace{-10pt}
	\end{figure}
	
	In this section we present a minimal solution for two-view relative motion estimation with known vertical direction, which uses only one affine correspondence, see Figure~\ref{fig:knownverticaldirection}. In this case, we have an IMU coupled with the camera. Assuming the roll and pitch angles of the camera can be obtained directly from the IMU, we can align every camera coordinate system with the measured gravity direction. The Y-axis of the camera is parallel to the gravity direction and the X-Z-plane of the camera is orthogonal to the gravity direction. The rotation matrix $\mathbf{R}_{\text{imu}}$ for aligning the camera coordinate system to the aligned camera coordinate system can be expressed:
	\vspace{-5pt}
	\begin{equation}
	\begin{aligned}
	&\mathbf{R}_{\text{imu}} = \mathbf{R}_{x}\mathbf{R}_{z}\\
	&= \begin{bmatrix}1&0&0\\
	0&\cos(\theta_x)&{\sin(\theta_x)}\\
	0&{-\sin(\theta_x)}&{\cos(\theta_x)}
	\end{bmatrix}\begin{bmatrix}
	{\cos(\theta_z)}&{\sin(\theta_z)}&0\\
	{ -\sin(\theta_z)}&{\cos(\theta_z)}&0\\
	0&0&1
	\end{bmatrix} \nonumber
	\end{aligned}
	\label{eq:RxRz}
	\end{equation}
	where $\theta_x$ and $\theta_z$ represent pitch and roll angle, respectively.
	
	Furthermore, denote $\mathbf{R}^{i}_{\text{imu}}$ and $\mathbf{R}^{j}_{\text{imu}}$ as the orientation information delivered by the IMU for views $i$ and $j$, respectively. Then the aligned image coordinates in views $i$ and $j$ can be expressed by
	\vspace{-6pt}
	\begin{equation}
	\mathbf{\tilde{p}}_{i} = \mathbf{R}^{i}_{\text{imu}}\mathbf{p}_{i},\qquad \mathbf{\tilde{p}}_{j} =  \mathbf{R}^{j}_{\text{imu}}\mathbf{p}_{j}.
	\label{eq:aligned image}
	\vspace{-6pt}
	\end{equation}
	
	By leveraging IMU measurement, the relative pose between original views $i$ and $j$ can be written as
	\vspace{-5pt}
	\begin{align}
	\begin{cases}
	\mathbf{R} =  (\mathbf{R}^{j}_{\text{imu}})^T \mathbf{R}_{y}\mathbf{R}^{i}_{\text{imu}}, \\
	\mathbf{t} =  (\mathbf{R}^{j}_{\text{imu}})^T \mathbf{\tilde{t}}.
	\end{cases}
	\label{eq:final relative pose}
	\vspace{-10pt}
	\end{align}
	where $\mathbf{R}_{y}$ is the rotation matrix between the aligned views $i$ and $j$, and $\mathbf{\tilde{t}}$ is the translation between the aligned views $i$ and $j$. Then the essential matrix between the original views $i$ and $j$ can be described as follows
	\vspace{-6pt}
	\begin{equation}
	\begin{aligned}
	\mathbf{E} &= [\mathbf{t}]_{\times}\mathbf{R} 
	= [(\mathbf{R}^{j}_{\text{imu}})^T \mathbf{\tilde{t}}]_{\times} (\mathbf{R}^{j}_{\text{imu}})^T \mathbf{R}_{y}\mathbf{R}^{i}_{\text{imu}}\\
	&=  (\mathbf{R}^{j}_{\text{imu}})^T\mathbf{\tilde{E}}\mathbf{R}^{i}_{\text{imu}}.
	\end{aligned}
	\label{eq:E_equ1}
	\end{equation}
	\vspace{-6pt}
	
	Note that $\mathbf{\tilde{E}}=[\mathbf{\tilde{t}}]_{\times} \mathbf{R}_{y}$ denotes the simplified essential matrix between the aligned views $i$ and $j$. Now, we substitute Eq.~\eqref{eq:E_equ1} into Eq.~\eqref{eq:E_Ac1}:
	\vspace{-2pt}
	\begin{equation}
	(({\mathbf{R}^{i}_{\text{imu}}})^T\mathbf{\tilde{E}}^T\mathbf{R}^{j}_{\text{imu}}\mathbf{p}_{j})_{(1:2)} =- (\mathbf{\hat{A}}^{T}(\mathbf{R}^{j}_{\text{imu}})^T\mathbf{\tilde{E}}\mathbf{R}^{i}_{\text{imu}} \mathbf{p}_{i})_{(1:2)}.
	\label{eq:E_Ac2}
	\end{equation}
	
	The above equation can be reformulated based on Eq.~\eqref{eq:aligned image}:
	\vspace{-4pt}
	\begin{equation}
	\begin{aligned}
	(({\mathbf{R}^{i}_{\text{imu}}})^T\mathbf{\tilde{E}}^T\mathbf{\tilde{p}}_{j})_{(1:2)} =  -(\mathbf{\hat{A}}^{T}(\mathbf{R}^{j}_{\text{imu}})^T\mathbf{\tilde{E}}\mathbf{\tilde{p}}_{i})_{(1:2)}.\\
	\end{aligned}
	\label{eq:E_Ac4}
	\vspace{-4pt}
	\end{equation}
	
	For further derivation, we denote $\mathbf{\tilde{p}}_{i}$, $\mathbf{\tilde{p}}_{j}$, $\mathbf{\tilde{E}}$ and $\mathbf{\tilde{A}}$ as follows
	\vspace{-4pt}
	{\small\begin{equation}
		\begin{aligned}
		&\mathbf{\tilde{p}}_{i}\triangleq[\tilde{u}_{i},\tilde{v}_{i},\tilde{w}_{i}]^{T}, \qquad	\mathbf{\tilde{p}}_{j}\triangleq[\tilde{u}_{j},\tilde{v}_{j},\tilde{w}_{j}]^{T} \\
		&\mathbf{\tilde{E}}=[\mathbf{\tilde{t}}]_{\times} \mathbf{R}_{y}
		= \begin{bmatrix}{0}&-\tilde{t}_z&\tilde{t}_y\\
		\tilde{t}_z&0&-\tilde{t}_x\\
		-\tilde{t}_y&\tilde{t}_x&0
		\end{bmatrix}\begin{bmatrix}{\cos{(\theta)}}&0&{-\sin{(\theta)}}\\
		0&1&0\\
		{ \sin{(\theta)}}&0&{\cos{(\theta)}}
		\end{bmatrix}\\
		&\ \ \ = \begin{bmatrix}{{{\tilde{t}}_y}\sin{(\theta)}}&-{\tilde{t}}_z&{\tilde{t}}_y{\cos{(\theta)}}\\
		{\tilde{t}}_z{\cos{(\theta)}}-{\tilde{t}}_x\sin{(\theta)}&0&-{\tilde{t}}_x\cos{(\theta)}-{\tilde{t}}_z{\sin{(\theta)}}\\
		-{\tilde{t}}_y{\cos{(\theta)}}&{\tilde{t}}_x&{\tilde{t}}_y{\sin{(\theta)}}
		\end{bmatrix}\\
		& \ \ \ \triangleq \begin{bmatrix}{e_1}&{e_2}&{e_3}\\
		{e_4}&0&{e_5}\\
		{-e_3}&{e_6}&{e_1}
		\end{bmatrix}\\
		&\mathbf{\tilde{A}} = \mathbf{\hat{A}}^{T}(\mathbf{R}^{j}_{\text{imu}})^T 
		\triangleq  \begin{bmatrix}{\tilde{a}_1}&{\tilde{a}_2}&{\tilde{a}_3}\\
		{\tilde{a}_4}&\tilde{a}_5&{\tilde{a}_6}\\
		{0}&{0}&{0}
		\end{bmatrix}\\
		&\mathbf{R}^{i}_{\text{imu}}= \mathbf{R}^{i}_{x}\mathbf{R}^{i}_{z} \triangleq  \begin{bmatrix}{\tilde{r}_1}&{\tilde{r}_2}&{0}\\
		{\tilde{r}_3}&\tilde{r}_4&{\tilde{r}_5}\\
		{\tilde{r}_6}&{\tilde{r}_7}&{\tilde{r}_8}
		\end{bmatrix}
		\end{aligned}
		\label{eq:E_Ac5}
		\vspace{-10pt}
		\end{equation}}
	
	By substituting Eq.~\eqref{eq:E_Ac5} into Eq.~\eqref{eq:E_Ac4}, we obtain two equations
	\vspace{-5pt}
	{\small\begin{align}
		&({\tilde{u}_i}{\tilde{a}_1} + {\tilde{w}_i}{\tilde{a}_3} + {\tilde{u}_j}{\tilde{r}_1} + {\tilde{w}_j}{\tilde{r}_6}){e_1} + ({\tilde{v}_i}{\tilde{a}_1} + {\tilde{u}_j}{\tilde{r}_3}){e_2} + \nonumber\\ 
		&({\tilde{w}_i}{\tilde{a}_1} + {\tilde{u}_j}{\tilde{r}_6} - {\tilde{u}_i}{\tilde{a}_3} - {\tilde{w}_j}{\tilde{r}_1}){e_3} + ({\tilde{u}_i}{\tilde{a}_2} + {\tilde{v}_j}{\tilde{r}_1}){e_4} + \nonumber\\ 
		& \qquad \qquad \qquad ({\tilde{w}_i}{\tilde{a}_2} + {\tilde{v}_j}{\tilde{r}_6}){e_5}  + ({\tilde{v}_i}{\tilde{a}_3} + {\tilde{w}_j}{\tilde{r}_3}){e_6}=0,
		\label{eq:Ac_eq1} \\
		&({\tilde{u}_i}{\tilde{a}_4} + {\tilde{w}_i}{\tilde{a}_6} + {\tilde{u}_j}{\tilde{r}_2} + {\tilde{w}_j}{\tilde{r}_7}){e_1} + ({\tilde{v}_i}{\tilde{a}_4} + {\tilde{u}_j}{\tilde{r}_4}){e_2}  + \nonumber\\
		&({\tilde{w}_i}{\tilde{a}_4} - {\tilde{u}_i}{\tilde{a}_6} + {\tilde{u}_j}{\tilde{r}_7} - {\tilde{w}_j}{\tilde{r}_2}){e_3} + ({\tilde{u}_i}{\tilde{a}_5} + {\tilde{v}_j}{\tilde{r}_2}){e_4} +  \nonumber\\
		& \qquad \qquad \qquad ({\tilde{w}_i}{\tilde{a}_5} + {\tilde{v}_j}{\tilde{r}_7}){e_5} + ({\tilde{v}_i}{\tilde{a}_6} + {\tilde{w}_j}{\tilde{r}_4}){e_6}=0. 
		\label{eq:Ac_eq2}
		\end{align}}
	
	\vspace{-15pt}
	In addition, the epipolar constraint $\mathbf{\tilde{p}}_{j}^T\mathbf{\tilde{E}}\mathbf{\tilde{p}}_{i} =0$ can be written as:
	\vspace{-4pt}
	{\small\begin{align}
		\begin{split}
		&({\tilde{u}_i}{\tilde{u}_j}+{\tilde{w}_i}{\tilde{w}_j}){e_1} + {\tilde{u}_j}{\tilde{v}_i}{e_2} + ({\tilde{u}_j}{\tilde{w}_i}-{\tilde{u}_i}{\tilde{w}_j}){e_3}\\
		& \qquad \qquad \qquad \qquad \ \ + {\tilde{u}_i}{\tilde{v}_j}{e_4}+{\tilde{v}_j}{\tilde{w}_i}{e_5}+{\tilde{v}_i}{\tilde{w}_j}{e_6}=0. \label{eq:Ac_eq3} \\
		\end{split}
		\vspace{-5pt}
		\end{align}}
	
	\vspace{-12pt}
	For an affine correspondence $(\mathbf{p}_{i},\mathbf{p}_{j},\mathbf{A})$, the combination of equations Eqs.~\eqref{eq:Ac_eq1}$\sim$\eqref{eq:Ac_eq3} can be expressed as $\mathbf{M}\mathbf{x} = \mathbf{0}$ , where $\mathbf{x} = [{e_1},\; {e_2},\; {e_3},\; {e_4},\; {e_5},\; {e_6}]^T$ is the vector of the unknown elements of the essential matrix. The null space of $\mathbf{M}$ is 3-dimensional. The solution of the polynomial equation system $\mathbf{x}$, which is up to a common scale, can be determined by the linear combination of three null space basis vectors: 
	\vspace{-9pt}
	\begin{equation}
	\begin{split}
	\mathbf{x} =  \beta\mathbf{m}_1+\gamma\mathbf{m}_2+\mathbf{m}_3,
	\label{eq:Ac_eq4} 
	\end{split}
	\end{equation}
	
	\vspace{-9pt}
	{\noindent}where the null space basis vectors $\{\mathbf{m}_i\}_{i=1,2,3}$ are computed from the SVD of matrix $\mathbf{M}$, and $\beta$ and $\gamma$ are the coefficients.  
	
	To determine the coefficients of $\beta$ and $\gamma$, note that there are two internal constraints for the essential matrix, \emph{i.e.}, the singularity of the essential matrix and the trace constraint:
	\vspace{-6pt}
	{\begin{align}
		&\qquad \qquad \determinat(\mathbf{\tilde{E}}) = 0,
		\label{eq:Ac_eq5} \\
		&2\mathbf{\tilde{E}}\mathbf{\tilde{E}}^T\mathbf{\tilde{E}} -\trace(\mathbf{\tilde{E}}\mathbf{\tilde{E}}^T)\mathbf{\tilde{E}}= 0.
		\label{eq:Ac_eq6}
		\vspace{-6pt}
		\end{align}}
	
	\vspace{-18pt}
	By substituting Eq.~\eqref{eq:Ac_eq4} into Eqs.~\eqref{eq:Ac_eq5} and~\eqref{eq:Ac_eq6}, a polynomial equation system with unknowns $\beta$ and $\gamma$ can be generated. A straightforward method to solve the equation system is using a general automatic solver generator~\cite{larsson2017efficient}. Inspired by \cite{fraundorfer2010minimal}, we use a more simpler method to convert the equation system to a univariate  quartic equation, see supplementary material for details. Once the coefficients $\beta$ and $\gamma$ have been obtained, the simplified essential matrix $\mathbf{\tilde{E}}$ is determined by Eq.~\eqref{eq:Ac_eq4} and can be decomposed into $\mathbf{R}_{y}$ and $\mathbf{\tilde{t}}$ by exploiting Eq.~\eqref{eq:E_Ac5}. Finally, the relative pose between views $i$ and $j$ can be obtained by Eq.~\eqref{eq:final relative pose}.
	
	%-------------------------------------------------------------------------
	\section{\label{sec:experiments}Experiments}
	The performance of the proposed methods is evaluated using both synthetic and real scene data. To deal with outliers, the minimal solvers can be integrated into a robust estimator using RANSAC or used for histogram voting. For the RANSAC, the relative pose which produces the highest number of inliers is chosen. For the histogram voting, we estimate the relative pose by selecting the peak of the histogram, which is formed by estimating poses from all the affine correspondences.
	
	%the performance of our methods is demonstrated on synthetic data in the presence of image noise and non-planar motion noise. 
	%using an affine correspondence based on epipolar geometry 
	For relative pose estimation under planar motion, the proposed solvers in Section~\ref{sec:1ACplane} are referred to as \texttt{1AC-Voting} (which uses histogram voting with the closed-form solution), \texttt{1AC-CS} (which uses RANSAC with the closed-form solution), and \texttt{1AC-LS} (which uses RANSAC with the least-squares solution). The solver for planar motion with unknown focal length in Section~\ref{sec:1ACUnkownF} is referred to as the \texttt{1AC-UnknownF}, which also uses RANSAC. The comparative methods include \texttt{6pt-Kukelova}\footnote{f+E+f relative pose solver.}~\cite{kukelova2017clever}, \texttt{5pt-Nister}~\cite{nister2004efficient}, \texttt{2AC-Barath}~\cite{barath2018efficient} and \texttt{2pt-Choi}~\cite{choi2018fast}. All comparative methods are integrated into a RANSAC scheme.
	
	%is demonstrated on synthetic data in the presence of image noise and IMU noise. The solver for relative pose estimation with known vertical direction
	For relative pose estimation with known vertical direction, our solver proposed in Section~\ref{sec:knownverticaldirection} is referred to as the \texttt{1AC method}. The proposed solver is compared against \texttt{5pt-Nister}~\cite{nister2004efficient}, \texttt{3pt-Sweeney}~\cite{sweeney2014solving}, \texttt{3pt-Saurer}~\cite{SaurerVasseur-457}, \texttt{2pt-Saurer}~\cite{SaurerVasseur-457} and \texttt{2AC-Barath}~\cite{barath2018efficient}. All of these minimal solvers are integrated into a RANSAC scheme. 
	
	Due to space limit, the efficiency comparison is provided in supplementary material. To demonstrate the suitability of our methods in real scenarios, the \texttt{KITTI} dataset~\cite{geiger2012Kitti} is used to validate the performance.%This experiment demonstrates that our methods are well suited for visual odometry in road driving scenarios, even though non-planar motion of vehicle exists.  
	
	%\vspace{-2pt}
	\subsection{Experiments on Synthetic Data}
	%\vspace{-2pt}
	%We evaluate the algorithms on synthetic data in the following setup. 
	The synthetic scene consists of a ground plane and 50 random planes, which are randomly distributed in the range of -5 to 5 meters (X-axis direction), -5 to 5 meters (Y-axis direction), and 10 to 20 meters (Z-axis direction). 50 points are randomly generated in the ground plane. We choose a point in each random plane randomly, so there are also 50 points in the random planes. The corresponding affine transformation related to each point correspondence is calculated from the homography, which is estimated by using four projected image points from the same plane~\cite{barath2019homography}. The baseline between two views is set to be 2 meters. The resolution of the camera is 640 $\times$ 480 pixels. The focal length is set to 400 pixels and the principal point is set to (320, 240).  
	
	The rotation and translation error are assessed by the root mean square error (RMSE) of the errors. We report the results on the data points within the first two intervals of a 5-quantile partitioning\footnote {k-quantiles divide an ordered dataset into $k$ regular intervals} (Quintile) of 1000 trials. The relative rotation and translation between views $i$ and $j$ are compared separately in the synthetic experiments. The rotation error compares the angular difference between the ground truth rotation and the estimated rotation. The translation error also compares the angular difference between the ground truth translation and the estimated translation since the estimated translation between views $i$ and $j$ is only known up to scale. Specifically, we define:
	\vspace{-3pt}
	\begin{itemize}
		\item Rotation error: ${\varepsilon _{\bf{R}}} = \arccos ((\trace({\mathbf{R}_{gt}}{{\mathbf{R}^T}}) - 1)/2)$
	\end{itemize}
	\vspace{-6pt}
	\begin{itemize}
		\item Translation error: ${\varepsilon _{\bf{t}}} = \arccos (({{\mathbf{t}_{gt}^T}}{\mathbf{t}})/(\left\| {\mathbf{t}_{gt}} \right\| \cdot \left\| {{\mathbf{t}}} \right\|))$
		\vspace{-3pt}
	\end{itemize}
	where $\mathbf{R}_{gt}$ and $\mathbf{t}_{gt}$ denote the ground truth rotation and translation, respectively. ${\mathbf{R}}$ and ${\mathbf{t}}$ denote the corresponding estimated rotation and translation, respectively.
	
	\vspace{-8pt}
	\subsubsection{Planar Motion Estimation}
\begin{figure}[tbp]
		\begin{center}
			\vspace{-8pt}
			\subfigure[\scriptsize{${\varepsilon _{\bf{R}}}$ with image noise}]
			{
				\includegraphics[width=0.47\linewidth]{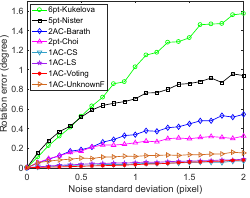}
			}
			\vspace{-10pt}
			\subfigure[\scriptsize{${\varepsilon _{\bf{t}}}$ with image noise}]
			{
				\includegraphics[width=0.463\linewidth]{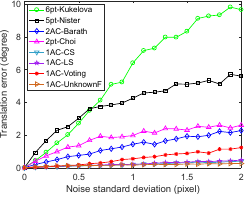}
			}
			\vspace{-8pt}
			\subfigure[\scriptsize{${\varepsilon _{\bf{R}}}$ with non-planar motion noise}]
			{
				\includegraphics[width=0.473\linewidth]{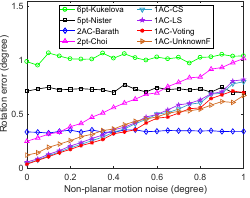}
			}
			\subfigure[\scriptsize{${\varepsilon _{\bf{t}}}$ with non-planar motion noise}]
			{
				\includegraphics[width=0.46\linewidth]{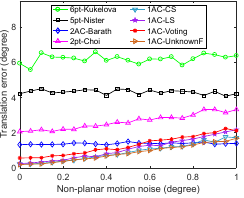}
			}
		\end{center}
		\caption{Rotation and translation error with planar motion estimation (unit: degree). (a)(b): vary image noise under perfect planar motion. (c)(d): vary non-planar motion noise and fix the image noise as $1.0$ pixel standard deviation. The left column reports the rotation error. The right column reports the translation error.}
		\label{fig:RT_planar}
		\vspace{-8pt}
	\end{figure}

	\vspace{-0pt}
	In this scenario the motion of the camera is described by ($\theta$, $\phi$), see Figure~\ref{fig:PlanarMotion}. Both angles vary from $-10^\circ$ to $10^\circ$.  Figure~\ref{fig:RT_planar}(a) and (b) show the performance of the proposed methods with respect to the magnitude of added image noise. All of our proposed methods for planar motion provide better results than comparative methods under perfect planar motion.
	It is worth to mention that our \texttt{1AC-UnknownF} method performs better than comparative methods even when the ground truth of the focal length is not used.	
	
	To test the performance of our method under non-planar motion, we generate the non-planar components of a 6DOF relative pose randomly and add them to the camera motion, which include X-axis rotation, Z-axis rotation, and direction of YZ-plane translation~\cite{choi2018fast}. The magnitude of non-planar motion noise is set to Gaussian noise with a standard deviation ranging from $0^\circ$ to $1^\circ$. The image noise is set to $1.0$ pixel standard deviation. Figure~\ref{fig:RT_planar}(c) and (d) show the performance of the proposed methods with respect to the magnitude of non-planar motion noise. The methods \texttt{6pt-Kukelova}, \texttt{5pt-Nister} and \texttt{2AC-Barath} do not have an obvious trend with non-planar motion noise levels, because these methods estimate 6DOF relative pose of two views. The proposed four methods perform better than the methods \texttt{6pt-Kukelova}, \texttt{5pt-Nister} and \texttt{2pt-Choi} at the maximum magnitude for the non-planar motion noise up to $1.0^\circ$. Meanwhile, the accuracy of these four methods is also better than the \texttt{2AC-Barath} method when the non-planar motion noise is less than $0.3^\circ$.
	
	\vspace{-10pt}
	\subsubsection{Motion with Known Vertical Direction}
	\vspace{-3pt}
	In this set of experiments the directions of the camera motion are set to forward, sideways and random motions, respectively. The second view is rotated around every axis, three rotation angles vary from $-10^\circ$ to $10^\circ$. The roll angle and pitch angle are known and used to align the camera coordinate system with the gravity direction. The proposed \texttt{1AC~method} is compared with \texttt{5pt-Nister}~\cite{nister2004efficient}, \texttt{3pt-Sweeney}~\cite{sweeney2014solving}, \texttt{3pt-Saurer}~\cite{SaurerVasseur-457},  \texttt{2pt-Saurer}~\cite{SaurerVasseur-457} and \texttt{2AC-Barath}~\cite{barath2018efficient}. Due to space
	limitations, we only show the results under random motion. The results under forward and sideways motions are available in the supplementary material. Figure~\ref{fig:RT_1AC}(a) and (b) show the performance of the proposed method with respect to the magnitude of image noise with perfect IMU data. Our method is robust to the increasing image noise and provides obviously better results than the previous methods. 
	
	\begin{figure}[tbp]
		\begin{center}
			\vspace{-4pt}
			\subfigure[\scriptsize{${\varepsilon _{\bf{R}}}$ with image noise}]
			{
				\includegraphics[width=0.47\linewidth]{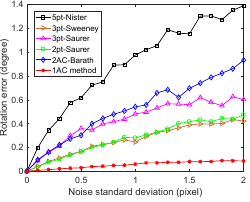}
			}
			\vspace{-10pt}
			\subfigure[\scriptsize{${\varepsilon _{\bf{t}}}$ with image noise}]
			{
				\includegraphics[width=0.46\linewidth]{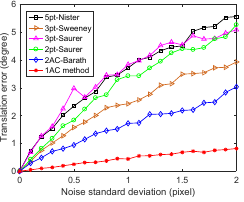}
			}
			\vspace{-10pt}
			\subfigure[\scriptsize{${\varepsilon _{\bf{R}}}$ with pitch angle noise}]
			{
				\includegraphics[width=0.473\linewidth]{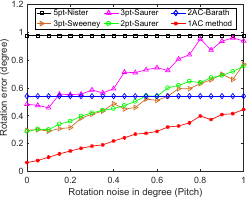}
			}
			\subfigure[\scriptsize{${\varepsilon _{\bf{t}}}$ with pitch angle noise}]
			{
				\includegraphics[width=0.46\linewidth]{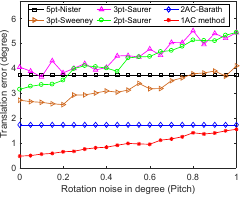}
			}
			\subfigure[\scriptsize{${\varepsilon _{\bf{R}}}$ with roll angle noise}]
			{
				\includegraphics[width=0.473\linewidth]{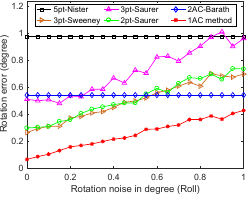}
			}
			\subfigure[\scriptsize{${\varepsilon _{\bf{t}}}$ with roll angle noise}]
			{
				\includegraphics[width=0.46\linewidth]{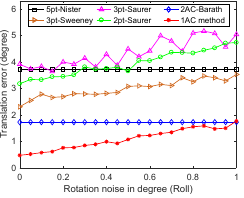}
			}
			\vspace{-8pt}
		\end{center}
		\caption{Rotation and translation error under random motion (unit: degree). (a)(b): vary image noise with perfect IMU data. (c)$\sim$(f): vary IMU angle noise and fix the image noise as $1.0$ pixel standard deviation. The left column reports the rotation error. The right column reports the translation error.}
		\label{fig:RT_1AC}
		\vspace{-5pt}
	\end{figure}
	
	Figure~\ref{fig:RT_1AC}(c)$\sim$(f) show the performance of the proposed method for increasing noise on the IMU data, while the image noise is set to $1.0$ pixel standard deviation. The \texttt{1AC~method} basically outperforms the methods \texttt{3pt-Sweeney}, \texttt{3pt-Saurer} and \texttt{2pt-Saurer}. The methods \texttt{5pt-Nister} and \texttt{2AC-Barath} are not influenced by the pitch error and the roll error, because their calculation does not utilize the known vertical direction as prior. It is interesting to see that our method performs better than the methods \texttt{5pt-Nister} and \texttt{2AC-Barath} in the random motion case, even though the rotation noise is around $1.0^\circ$. Under forward and sideways motion, 
	the accuracy of our method is also better than the methods \texttt{5pt-Nister} and \texttt{2AC-Barath}, when the rotation noise stays below $0.3^\circ$. 
	
	%\vspace{-4pt}
	\subsection{Experiments on Real Data}
	The performance of our methods on real image data is evaluated on the \texttt{KITTI} dataset~\cite{geiger2012Kitti}. All the sequences which provide ground truth data are utilized in this experiments. There are about 23000 images in total and are available as sequence 0 to 10.
	%\subsubsection{Comparison of rotation and translation estimation to ground truth}
	\vspace{-8pt}
	\subsubsection{Pose Estimation from Pairwise Image Pairs}
	Two settings of experiments are performed with the \texttt{KITTI} dataset, including planar motion estimation and relative pose estimation with known vertical direction. The ASIFT feature extraction and matching~\cite{morel2009asift} is performed to obtain the affine correspondences between consecutive frames. Both the histogram voting and the RANSAC schemes are tested in this experiment. An inlier threshold of $2$ pixels and a fixed number of $100$ iterations are set in RANSAC.
	
	In the first experiment, we test the relative pose estimation algorithms under planar motion. The motion estimation results between two consecutive images ($\theta$, $\phi$) are compared to the corresponding ground truth. 
	The median error for each individual sequence is used to evaluate the performance. The proposed methods are compared with \texttt{2pt-Choi}~\cite{choi2018fast}. 
	The results of the rotation and translation error under planar motion assumption are shown in Table~\ref{PlaneRTErrror}.  Table~\ref{PlaneRTErrror} demonstrates that all of our planar motion methods provide better results than the \texttt{2pt-Choi} method. The overall performance of the \texttt{1AC-Voting} method is best among all the methods, particularly the rotation accuracy of the \texttt{1AC-Voting} method is significantly high than other methods.
	
	\begin{table}[htbp]
		\vspace{-5pt}
		\begin{center}
			\setlength{\tabcolsep}{0.8mm}{
				\scalebox{0.975}{
					\begin{tabular}{|c||c|c|c|c|}
						\hline
						\multirow{2}{*}{\footnotesize{Seq.}}  &  \small{2pt}\scriptsize{-Choi~\cite{choi2018fast}} & \textbf{\small{1AC-CS}}   &  \textbf{\small{1AC-LS}} & \textbf{\small{1AC-Voting}}    \\
						\cline{2-5}
						&   ${\varepsilon _{\bf{R}}}$\qquad\ ${\varepsilon _{\bf{t}}}$   &   ${\varepsilon _{\bf{R}}}$\qquad\ ${\varepsilon _{\bf{t}}}$ &   ${\varepsilon _{\bf{R}}}$\qquad\ ${\varepsilon _{\bf{t}}}$  &   ${\varepsilon _{\bf{R}}}$\qquad\ ${\varepsilon _{\bf{t}}}$ \\
						\hline
						00                  &  0.203 \  5.169&  0.133  \ \textbf{1.335}&  0.155  \          1.345&    \textbf{0.016} \          1.493\\
						\rowcolor{gray!10}01&  0.150 \  3.617&  0.117  \ \textbf{1.135}&  0.134  \          1.149&    \textbf{0.010} \          1.165\\
						02                  &  0.154 \  3.364&  0.062  \          1.152&  0.082  \          1.191&    \textbf{0.017} \ \textbf{1.029}\\
						\rowcolor{gray!10}03&  0.177 \  6.441&  0.084  \          1.157&  0.100  \ \textbf{1.152}&    \textbf{0.013} \          1.225\\
						04                  &  0.115 \  2.871&  0.029  \          1.132&  0.041  \          1.155&    \textbf{0.012} \ \textbf{1.018}\\
						\rowcolor{gray!10}05&  0.143 \  4.407&  0.071  \ \textbf{1.276}&  0.085  \          1.304&    \textbf{0.011} \          1.614\\
						06                  &  0.152 \  3.379&  0.051  \ \textbf{1.302}&  0.068  \          1.340&    \textbf{0.008} \          1.655\\
						\rowcolor{gray!10}07&  0.127 \  4.764&  0.059  \          1.487&  0.074  \ \textbf{1.462}&    \textbf{0.014} \          1.769\\
						08                  &  0.137 \  4.312&  0.064  \          1.428&  0.081  \ \textbf{1.427}&    \textbf{0.014} \          1.591\\
						\rowcolor{gray!10}09&  0.141 \  3.508&  0.062  \ \textbf{1.215}&  0.081  \          1.218&    \textbf{0.021} \          1.221\\
						10                  &  0.145 \  3.829&  0.067  \ \textbf{1.299}&  0.090  \ \textbf{1.299}&    \textbf{0.018} \          1.464\\
						\hline
			\end{tabular}}}
		\end{center}
		\vspace{-5pt}
		\caption{Rotation and translation error for \texttt{KITTI} sequences under planar motion assumption (unit: degree).}
		\label{PlaneRTErrror}
	\end{table}	
	
	\vspace{-20pt}
	In the second experiment, we test the relative pose estimation algorithm with known vertical direction, \emph{i.e.}, \texttt{1AC method}. To simulate IMU measurements which provide a known gravity vector for the views of the camera, the image coordinates are pre-rotated by ${\mathbf{R}_x}{\mathbf{R}_z}$ obtained from the ground truth data. Table~\ref{VerticalRTErrror} lists the results of the rotation and translation estimation. The proposed methods are also compared against \texttt{5pt-Nister}~\cite{nister2004efficient}, \texttt{3pt-Sweeney}~\cite{sweeney2014solving}, \texttt{3pt-Saurer}~\cite{SaurerVasseur-457}, \texttt{2pt-Saurer}~\cite{SaurerVasseur-457} and \texttt{2AC-Barath}~\cite{barath2018efficient}. Table~\ref{VerticalRTErrror} demonstrates that our method is significantly more accurate than the other methods, except for the translation error of sequences 02, 09 and 10.
	%\vspace{-5pt}
	\begin{table}[t]
		\begin{center}
			\setlength{\tabcolsep}{0.8mm}{
				\scalebox{0.71}{
					\begin{tabular}{|c||c|c|c|c|c|c|}  
						\hline
						\multirow{2}{*}{\footnotesize{Seq.}} &  \small{5pt}\scriptsize{-Nister~\cite{nister2004efficient}} &  \small{3pt}\scriptsize{-Sweeney~\cite{sweeney2014solving}} &  \small{3pt}\scriptsize{-Saurer~\cite{SaurerVasseur-457}}  & \small{2pt}\scriptsize{-Saurer~\cite{SaurerVasseur-457}}  &  \small{2AC}\scriptsize{-Barath~\cite{barath2018efficient}}& \textbf{\small{1AC method}} \\
						\cline{2-7}
						& ${\varepsilon _{\bf{R}}}$\qquad\ ${\varepsilon _{\bf{t}}}$      &  ${\varepsilon _{\bf{R}}}$\qquad\ ${\varepsilon _{\bf{t}}}$      &   ${\varepsilon _{\bf{R}}}$\qquad\ ${\varepsilon _{\bf{t}}}$     &   ${\varepsilon _{\bf{R}}}$\qquad\ ${\varepsilon _{\bf{t}}}$     &   ${\varepsilon _{\bf{R}}}$\qquad\ ${\varepsilon _{\bf{t}}}$ &   ${\varepsilon _{\bf{R}}}$\qquad\ ${\varepsilon _{\bf{t}}}$\\
						\hline
						00                  &  .137 \  2.254&.065  \  2.165   &.153  \          2.231&  .336   \  7.675&.196  \   4.673&  \textbf{.038}  \  \textbf{2.006}  \\
						\rowcolor{gray!10}01&  .120 \  1.988&.082  \  2.342   &.091  \          2.211&  .186   \  9.806&.111  \   4.198&  \textbf{.050}  \  \textbf{1.507}  \\
						02                  &  .134 \  1.787&.059  \ \textbf{1.658} &.113 \  1.723&  .293   \  6.034&.251  \   4.694&  \textbf{.039}  \           1.861  \\      
						\rowcolor{gray!10}03&  .109 \  2.507&.067  \  2.723   &.161  \          2.620&  .316   \  9.249&.175  \   6.064&  \textbf{.041}  \  \textbf{2.143}  \\
						04                  &  .111 \  1.692&.048  \  1.558   &.043  \          1.616&  .141   \  4.816&.184  \   4.036&  \textbf{.033}  \  \textbf{1.538}  \\
						\rowcolor{gray!10}05&  .116 \  2.059&.054  \  1.895   &.115  \          1.961&  .253   \  7.238&.162  \   4.481&  \textbf{.031}  \  \textbf{1.725}  \\
						06                  &  .130 \  1.783&.068  \  1.615   &.111  \          1.658&  .232   \  5.750&.176  \   4.026&  \textbf{.046}  \  \textbf{1.538}  \\
						\rowcolor{gray!10}07&  .113 \  2.434&.052  \  2.183   &.159  \          2.217&  .378   \  8.293&.161  \   4.649&  \textbf{.033}  \  \textbf{2.009}  \\
						08                  &  .122 \  2.335&.053  \  2.216   &.102  \          2.266&  .241   \  7.556&.182  \   5.044&  \textbf{.036}  \  \textbf{2.201}  \\
						\rowcolor{gray!10}09&  .133 \  1.843&.059  \  \textbf{1.701}   &.176  \          1.812&  .409   \  6.606&.224  \   4.924&  \textbf{.045}  \  1.799  \\
						10                  &  .131 \  1.839&.059  \  \textbf{1.750}   &.145  \          2.004&  .308   \  7.324&.216  \   4.520&  \textbf{.037}  \          1.935  \\
						\hline
			\end{tabular}}}
		\end{center}
		\vspace{-5pt}
		\caption{Rotation and translation error for \texttt{KITTI} sequences with known vertical direction (unit: degree).}
		\label{VerticalRTErrror}
		\vspace{-14pt}
	\end{table}
	
	\vspace{-14pt}
	\subsubsection{Visual Odometry}
	\vspace{-4pt} 
	We demonstrate the usage of the \texttt{1AC~method} in a monocular visual odometry pipeline to evaluate its performance in a real application. Our monocular visual odometry is based on \texttt{ORB-SLAM2}~\cite{mur2017orb}. The affine correspondences extracted by ASIFT feature matching are used to replace the ORB features. The relative pose between two consecutive frames is estimated based on the combination of the \texttt{1AC~method} using RANSAC, and is used to replace the original map initialization and the constant velocity motion model. The estimated trajectories after alignment with ground truth are illustrated in Figure~\ref{fig:trajectory}. The color along the trajectory encodes the absolute trajectory error (ATE)~\cite{sturm2012benchmark}. Due to space limit, we show the trajectories of two sequences only. The results of other sequences can be found in supplementary materials\footnote{Both \texttt{ORB-SLAM2} and our monocular visual odometry fail to produce a valid result for sequence 01, because it is a highway with few tractable close objects.}. It can be seen that the proposed \texttt{1AC~method} method has the smallest ATE among the compared trajectories. %For more results see the supplementary material.	         
	\begin{figure}[htbp]
		%	\vspace{-5pt}
		\begin{center}
			\subfigure[Seq.00]
			{
				\includegraphics[width=1.0\linewidth]{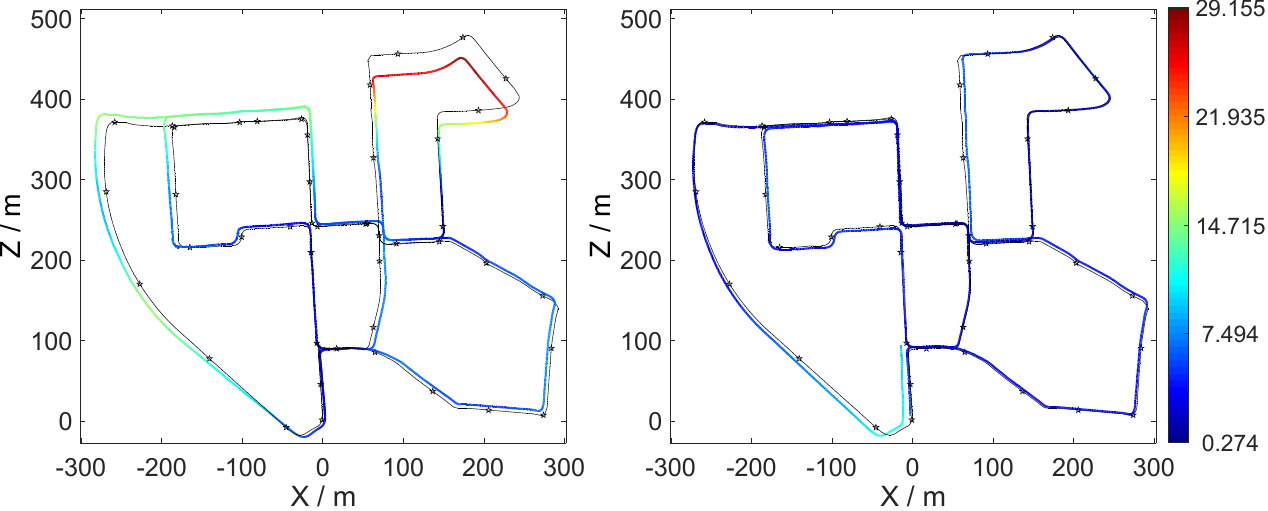}
			}
			\vspace{-6pt}
			\subfigure[Seq.02]
			{
				\includegraphics[width=1.0\linewidth]{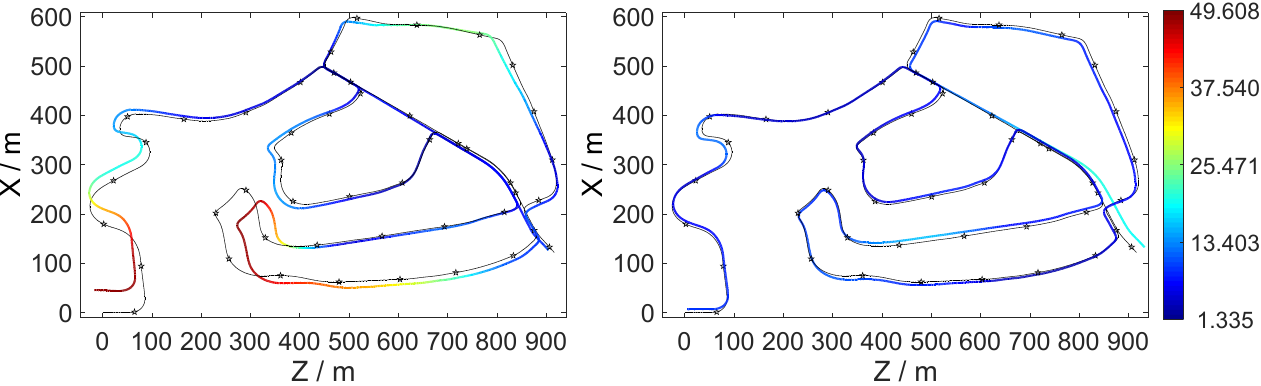}
			}
			\vspace{-6pt}
		\end{center}
		\caption{Estimated visual odometry trajectories. The left column reports the results of \texttt{ORB-SLAM2}. The right column reports the results of our monocular visual odometry. Colorful curves are estimated trajectories, and black curves with stars are ground truth trajectories. Best viewed in color.}
		\label{fig:trajectory}
		\vspace{-0pt}
	\end{figure}
	
	\begin{table}[htbp]
		\begin{center}
			\setlength{\tabcolsep}{0.8mm}{
				\scalebox{0.845}{
					\begin{tabular}{|c||c|c|c||c|c|}
						\hline
						\multirow{2}{*}{\footnotesize{Seq.}} &  \small{ORB}\scriptsize{-SLAM2~\cite{mur2017orb}} & \textbf{\small{1AC}\scriptsize{-SLAM}} & \multirow{2}{*}{\footnotesize{Seq.}} &  \small{ORB}\scriptsize{-SLAM2~\cite{mur2017orb}} & \textbf{\small{1AC}\scriptsize{-SLAM}}\\
						\cline{2-3} \cline{5-6}
						&  ${\varepsilon _{\bf{R}}}$\qquad\ ${\varepsilon _{\bf{t}}}$   & ${\varepsilon _{\bf{R}}}$\qquad\ ${\varepsilon _{\bf{t}}}$  &   &  ${\varepsilon _{\bf{R}}}$\qquad\ ${\varepsilon _{\bf{t}}}$   & ${\varepsilon _{\bf{R}}}$\qquad\ ${\varepsilon _{\bf{t}}}$\\
						\hline
						00&         0.821  \ 0.923&  \textbf{0.803}  \   \textbf{0.421}&06&  0.142          \           1.478&  \textbf{0.126}  \   \textbf{0.995}\\
						\rowcolor{gray!10}02&         0.200  \ 1.052&  \textbf{0.156}  \   \textbf{0.686}&07&  0.149          \           0.879&  \textbf{0.137}  \   \textbf{0.330}\\
						03& \textbf{0.113} \ 0.244&         {0.118}  \   \textbf{0.185}&08&  0.177          \           1.778&  \textbf{0.159}  \   \textbf{0.659}\\
						\rowcolor{gray!10}04&         0.151  \ 0.417&  \textbf{0.097}  \   \textbf{0.307}&09&  0.221          \           0.777&  \textbf{0.172}  \   \textbf{0.502}\\
						05&         0.264  \ 0.681&  \textbf{0.254}  \   \textbf{0.306}&10&  \textbf{0.129} \  \textbf{0.633}&        0.238     \           1.008\\
						\hline
			\end{tabular}}}
		\end{center}
		\caption{RMSE for rotation and translation using the RPE metric for \texttt{KITTI} sequences. Rotation error unit: degree. Translation error unit: meter.}
		\label{RPE}
		\vspace{-14pt}
	\end{table}
	
	Moreover, we also evaluate the Relative Pose Error (RPE) between the estimated trajectory and the ground truth trajectory, which measures the relative accuracy of the trajectory over fixed time intervals~\cite{sturm2012benchmark}. The RMSE for rotation and translation using the RPE metric is illustrated in Table.~\ref{RPE}. %\footnote {ORB-SLAM2 fails to produce a valid results for Seq. 01 due to few trackable close objects}. 
	Our monocular visual odometry generally has smaller rotation and translational errors than \texttt{ORB-SLAM2}. 
	
	\vspace{4pt} 	      	
	\section{\label{sec:conclusion}Conclusion}
	\vspace{-2pt} 
	In this paper, we showed that by exploiting the affine parameters it is possible to estimate the relative pose of a camera with only one affine correspondence under the planar motion assumption. Three minimal case solutions have been proposed to recover the planar motion of camera, amongst which is a solver which can even deal with an unknown focal length. In addition, a minimal case solution has been proposed to estimate the relative pose of a camera for the case of a known vertical direction. The assumptions in these methods are common to scenes in which self-driving cars and ground robots operate. By evaluating our algorithms on synthetic data and real-world image data sets, we demonstrate that our method can be used efficiently for outlier removal and for initial motion estimation in visual odometry.\\	
	
	\vspace{-20pt} 	
	\section*{Acknowledgments}
	\vspace{-3pt} 
	This work has been partially funded by the National Natural Science Foundation of China (11902349).
	
	\appendix
	\large
	\begin{center}
		{\bf Supplementary Material }
	\end{center}
	\normalsize

	%%%%%%%%% BODY TEXT
	\section{\label{sec:LeastSquaresSolution}Least-Squares Solution}
	Refer to Eq.~\eqref{equ:LagMul1} in the paper, by taking the partial derivatives with $\{x_i\}_{i=1}^4$ and $\{\lambda_i\}_{i=1}^2$ and set them to be zeros, we obtain an equation system with unknowns $\{x_i\}_{i=1}^4$ and $\{\lambda_i\}_{i=1}^2$:
	\vspace{-2pt}
	{\small \begin{align}
	\frac{1}{2}\frac{\partial{L}}{\partial x_1} &= \sum_{i=1}^3 \left[ {a_i^2}x_1 + a_i (b_i x_2 + c_i x_3 + d_i x_4) \right] + \lambda_1 x_1 = 0 \nonumber \\
	\frac{1}{2}\frac{\partial{L}}{\partial x_2} &= \sum_{i=1}^3 \left[ {b_i^2} x_2 + b_i (a_i x_1 + c_i x_3 + d_i x_4) \right] + \lambda_1 x_2 = 0 \nonumber \\
	\frac{1}{2}\frac{\partial{L}}{\partial x_3} &= \sum_{i=1}^3 \left[ {c_i^2}x_3 + c_i (a_i x_1 + b_i x_2 + d_i x_4) \right] + \lambda_2 x_3 = 0 \nonumber \\
	\frac{1}{2}\frac{\partial{L}}{\partial x_4} &= \sum_{i=1}^3 \left[ {d_i^2} x_4 + d_i (a_i x_1 + b_i x_2 + c_i x_3) \right] + \lambda_2 x_4 = 0 \nonumber \\
	\frac{\partial{L}}{\partial \lambda_1} &= x_1^2 + x_2^2 - 1 = 0 \nonumber \\
	\frac{\partial{L}}{\partial \lambda_2} &= x_3^2 + x_4^2 - 1 = 0 \nonumber
	\end{align}}
	
	\vspace{-9pt}
	The above equation system contains $6$ unknowns $\{x_1, x_2, x_3, x_4, \lambda_1, \lambda_2\}$, and the order is $2$. 
	
	%-------------------------------------------------------------------------
	\section{\label{sec:knownverticaldirection}Relative Pose Estimation with Known Vertical Direction}
	We show the solution procedure of the coefficients $\beta$ and $\gamma$. To derive the solution, we start by substituting Eq.~\eqref{eq:Ac_eq4} into Eqs.~\eqref{eq:Ac_eq5} and~\eqref{eq:Ac_eq6} in the paper. Six equations from the trace constraint Eq.~\eqref{eq:Ac_eq6}, together with a equation from the singularity of the essential matrix Eq.~\eqref{eq:Ac_eq5}, form a system of $7$ polynomial equations in $2$ unknowns $\{\beta, \gamma\}$, which has a maximum polynomial degree of $3$. First, we stack $7$ polynomial equations into a matrix form as
	\vspace{-4pt}
	\begin{equation}
	\mathbf{M}_1\mathbf{v}_1 = 0,
	\label{eq:equations set1}
	\end{equation}
	where $\mathbf{v}_1=[{\beta}^3, {\beta}^2{\gamma}, {\beta}^2, {\beta}{\gamma}^2, {\beta}{\gamma}, {\beta}, {\gamma}^3, {\gamma}^2, {\gamma}, 1]^T$, $\mathbf{M}_1$ is a 7$\times$10 coefficient matrix. 
	
	Since there is a linear dependency between the elements of the essential matrix, \emph{i.e.}, ${e_2}$, ${e_4}$, ${e_5}$ and ${e_6}$, the rank of the coefficient matrix $\mathbf{M}_1$ is only 6. By performing Gaussian elimination and row operations on the $6$ linearly independent equations, we set up a new polynomial equation system as follows:
	\vspace{-4pt}
	\[
	\renewcommand\arraystretch{1.3}
	%\mleft[
	\begin{array}{c c c c c c c c c c c}
	{\beta}^3 & {\beta}^2{\gamma} &{\beta}^2 & {\beta}{\gamma}^2 & {\beta}{\gamma} & {\beta} & {\gamma}^3 & {\gamma}^2 & {\gamma} & 1 \\
	\hline
	1 &  &  &  &  &  &. &. &. &. &      \\
	&1 &  &  &  &  &. &. &. &. &      \\
	&  &1 &  &  &  &. &. &. &. &      \\
	&  &  &1 &  &  &. &. &. &. &      \\
	&  &  &  &1 &  &. &. &. &. &\left \langle Q_a \right \rangle \\
	&  &  &  &  &1 &. &. &. &. &\left \langle Q_b \right \rangle \\    
	\end{array}
	\label{eq:GaussianElimination}
	%\mright]
	\]
	where ${Q_a}=poly({\beta}{\gamma},{\gamma}^3,{\gamma}^2,{\gamma},1)$ and ${Q_b}=ploy({\beta}, {\gamma}^3,{\gamma}^2,{\gamma},1)$ represent the polynomial in the fifth and sixth rows, respectively. 
	
	In order to eliminate the monomial ${\beta}{\gamma}$, we multiply ${Q_b}$ with ${\gamma}$ and subtract it from ${Q_a}$:
	\begin{equation}
	{Q_c} = {\gamma}{Q_b} - {Q_a} = poly({\gamma}^4,{\gamma}^3,{\gamma}^2,{\gamma},1)
	\label{eq:equationForgamma}
	\end{equation}
	
	Now, we get an up to degree 4 polynomial in ${\gamma}$: ${Q_c}$. The unknown ${\gamma}$ has at most 4 solutions and can be computed as the eigenvalues of the companion matrix of ${Q_c}$. Then the corresponding solution for the unknown ${\beta}$ is obtained directly by substituting ${\gamma}$ into ${Q_b}$.
	
	%-------------------------------------------------------------------------
	\section{\label{sec:experiments}Experiments}
	
	\subsection{Efficiency Comparison}
	We evaluate the run-times of our solvers and the comparative solvers on an Intel(R) Core(TM) i7-8550U 1.80GHz using MATLAB. All algorithms are implemented in Matlab, except that the \texttt{5pt-Nister} method is implemented in C by using mex file. All timings are averaged over 10000 runs. Table~\ref{PlaneTime} summarizes the run-times for the planar motion estimation algorithms\footnote {Note that the run-times of the methods \texttt{5pt-Nister} and \texttt{2AC-Barath} are showed in Table~\ref{VerticalTime}.}. The run-times of the methods \texttt{1AC-Voting} and \texttt{1AC-CS} are same and quite low, because both methods use the same solver and the computational complexity is mainly about computing the eigenvector of the matrix. For the methods \texttt{1AC-LS} and \texttt{1AC-UnknownF}, the high run-times are due to the complexity of the Gr\"{o}bner basis solution.              
	\vspace{-2pt}
	\begin{table}[htbp]
		\begin{center}
			\setlength{\tabcolsep}{0.8mm}{
				\scalebox{0.716}{
					\begin{tabular}{|c||c|c|c|c|c|c|}
						\hline
						\small{Methods}  &  \small{6pt}\scriptsize{-Kukelova~\cite{kukelova2017clever}} &  \small{2pt}\scriptsize{-Choi~\cite{choi2018fast}} & \textbf{\small{1AC-CS}}   &  \textbf{\small{1AC-LS}} & \textbf{\small{1AC-Voting}} & \textbf{\small{1AC-UnknownF}}    \\
						\hline
						\small{Timings}& 0.405 &  0.098&	\textbf{0.007}&	0.120&	\textbf{0.007}& 0.196\\ 			
						\hline
			\end{tabular}}}
		\end{center}
	    \vspace{-1pt}
		\caption{Run-time comparison of planar motion estimation algorithms (unit: $ms$).}
		\label{PlaneTime}
	\end{table}	
	
	\vspace{-4pt}
	Table~\ref{VerticalTime} summarizes the run-times for the motion estimation algorithms with known vertical direction. The run-time of the \texttt{3pt-Saurer} method is higher than the \texttt{1AC method} method due to the complexity of the Gr\"{o}bner basis solution. Since the mex file is used, the run-time of the \texttt{5pt-Nister} method is low. The run-time of the \texttt{1AC method} method is significantly lower than the \texttt{2AC-Barath} method, because the essential matrix between two views is simplified when the common direction of rotation is known, and we use a low-complexity approach to solve the essential matrix as shown in Section~\ref{sec:knownverticaldirection}.    
	\begin{table}[htbp]
		\begin{center}
			\setlength{\tabcolsep}{0.8mm}{
				\scalebox{0.672}{
					\begin{tabular}{|c||c|c|c|c|c|c|}
						\hline
						\small{Methods} &  \small{5pt}\scriptsize{-Nister~\cite{nister2004efficient}} &  \small{3pt}\scriptsize{-Sweeney~\cite{sweeney2014solving}}
						&  \small{3pt}\scriptsize{-Saurer~\cite{SaurerVasseur-457}}  & \small{2pt}\scriptsize{-Saurer~\cite{SaurerVasseur-457}}  &  \small{2AC}\scriptsize{-Barath~\cite{barath2018efficient}}& \textbf{\small{1AC method}} \\
						\hline
						\small{Timings}&  0.118& 0.174 & 2.066&	\textbf{0.097}&	65.101&	1.212\\
						\hline
			\end{tabular}}}
		\end{center}
		\vspace{-1pt}
		\caption{Run-time comparison of motion estimation algorithms with known vertical direction (unit: $ms$).}
		\label{VerticalTime}
	\end{table}
	
	\subsection{Motion with Known Vertical Direction}
\begin{figure}[htbp]
		\begin{center}
			\subfigure[\scriptsize{${\varepsilon _{\bf{R}}}$ with image noise}]
			{
				\includegraphics[width=0.47\linewidth]{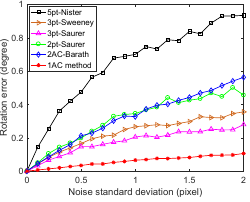}
			}
			\subfigure[\scriptsize{${\varepsilon _{\bf{t}}}$ with image noise}]
			{
				\includegraphics[width=0.46\linewidth]{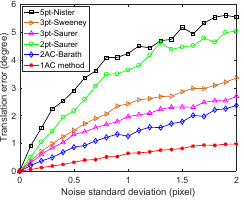}
			}
			\subfigure[\scriptsize{${\varepsilon _{\bf{R}}}$ with pitch angle noise}]
			{
				\includegraphics[width=0.47\linewidth]{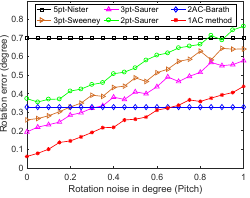}
			}
			\subfigure[\scriptsize{${\varepsilon _{\bf{t}}}$ with pitch angle noise}]
			{
				\includegraphics[width=0.46\linewidth]{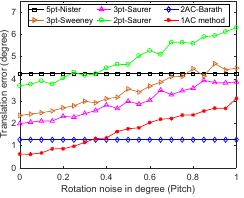}
			}
			\subfigure[\scriptsize{${\varepsilon _{\bf{R}}}$ with roll angle noise}]
			{
				\includegraphics[width=0.47\linewidth]{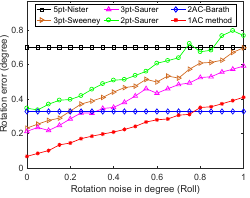}
			}
			\subfigure[\scriptsize{${\varepsilon _{\bf{t}}}$ with roll angle noise}]
			{
				\includegraphics[width=0.46\linewidth]{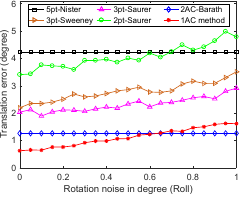}
			}
		\end{center}
		\vspace{-8pt}
		\caption{Rotation and translation error under forward motion (unit: degree). (a)(b): vary image noise with perfect IMU data. (c)$\sim$(f): vary IMU angle noise and fix the image noise as $1.0$ pixel standard deviation. The left column reports the rotation error. The right column reports the translation error.}
		\label{fig:RTForwardMotion_1AC}
		%\vspace{-6pt}
	\end{figure}

\begin{figure}[htbp]
        \vspace{-14pt}
		\begin{center}
			\subfigure[\scriptsize{${\varepsilon _{\bf{R}}}$ with image noise}]
			{
				\includegraphics[width=0.47\linewidth]{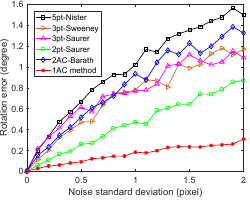}
			}
			\vspace{-10pt}
			\subfigure[\scriptsize{${\varepsilon _{\bf{t}}}$ with image noise}]
			{
				\includegraphics[width=0.46\linewidth]{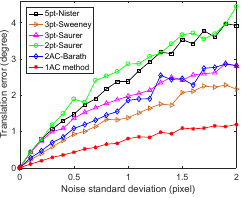}
			}
			\vspace{-10pt}
			\subfigure[\scriptsize{${\varepsilon _{\bf{R}}}$ with pitch angle noise}]
			{
				\includegraphics[width=0.47\linewidth]{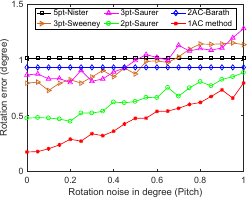}
			}
			\subfigure[\scriptsize{${\varepsilon _{\bf{t}}}$ with pitch angle noise}]
			{
				\includegraphics[width=0.46\linewidth]{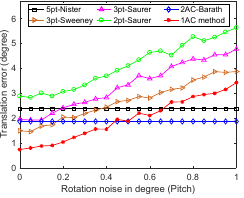}
			}
			\vspace{-10pt}
			\subfigure[\scriptsize{${\varepsilon _{\bf{R}}}$ with roll angle noise}]
			{
				\includegraphics[width=0.47\linewidth]{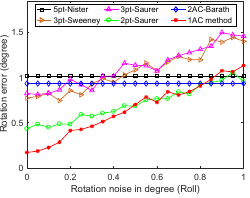}
			}
			\subfigure[\scriptsize{${\varepsilon _{\bf{t}}}$ with roll angle noise}]
			{
				\includegraphics[width=0.465\linewidth]{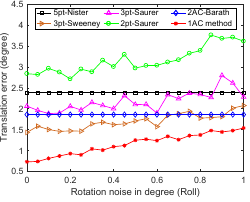}
			}
		\end{center}
		\caption{Rotation and translation error under sideways motion (unit: degree). (a)(b): vary image noise with perfect IMU data. (c)$\sim$(f): vary IMU angle noise and fix the image noise as $1.0$ pixel standard deviation. The left column reports the rotation error. The right column reports the translation error.}
		\label{fig:RTSidewaysMotion_1AC}
	\end{figure}

	In this section we show the performance of the proposed \texttt{1AC~method} under forward and sideways motion. Figure~\ref{fig:RTForwardMotion_1AC} shows the performance of the proposed method under forward motion. Figure~\ref{fig:RTSidewaysMotion_1AC} shows the performance of the proposed method under sideways motion.
	
	\subsection{Visual Odometry}
	Here we show more trajectories for the experiments with \texttt{KITTI} dataset\footnote {Both \texttt{ORB-SLAM2} and our monocular visual odometry fail to produce a valid result for sequence 01, because it is a highway with few tractable close objects.}, see Figure~\ref{fig:trajectory}. It shows that the proposed \texttt{1AC~method} method has the smallest ATE among all the compared trajectories.
	\begin{figure}[htbp]
		\begin{center}
			\subfigure[Seq.03]
			{
				\includegraphics[width=1.0\linewidth]{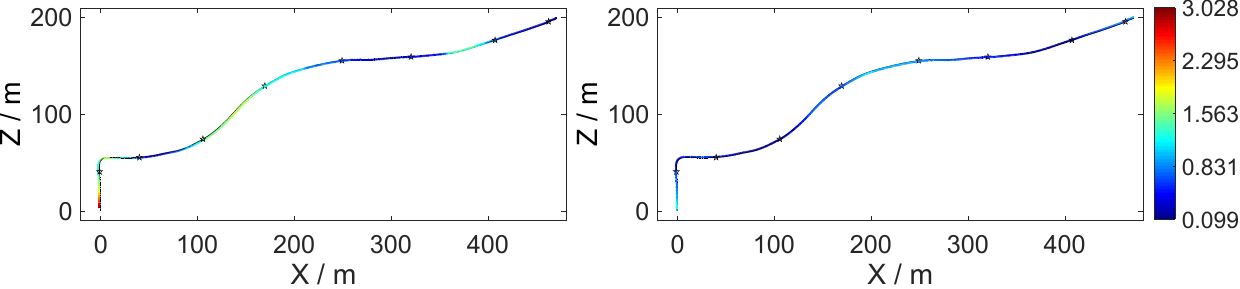}
			}
		    \vspace{-10pt}
			\subfigure[Seq.04]
			{
				\includegraphics[width=1.0\linewidth]{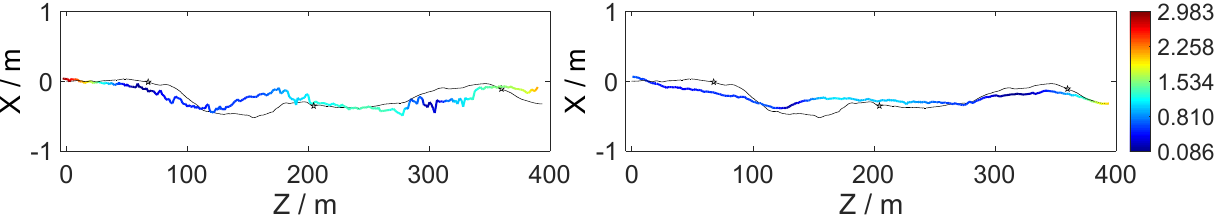}
			}
			\vspace{-10pt}
            \subfigure[Seq.05]
            {
            	\includegraphics[width=1.0\linewidth]{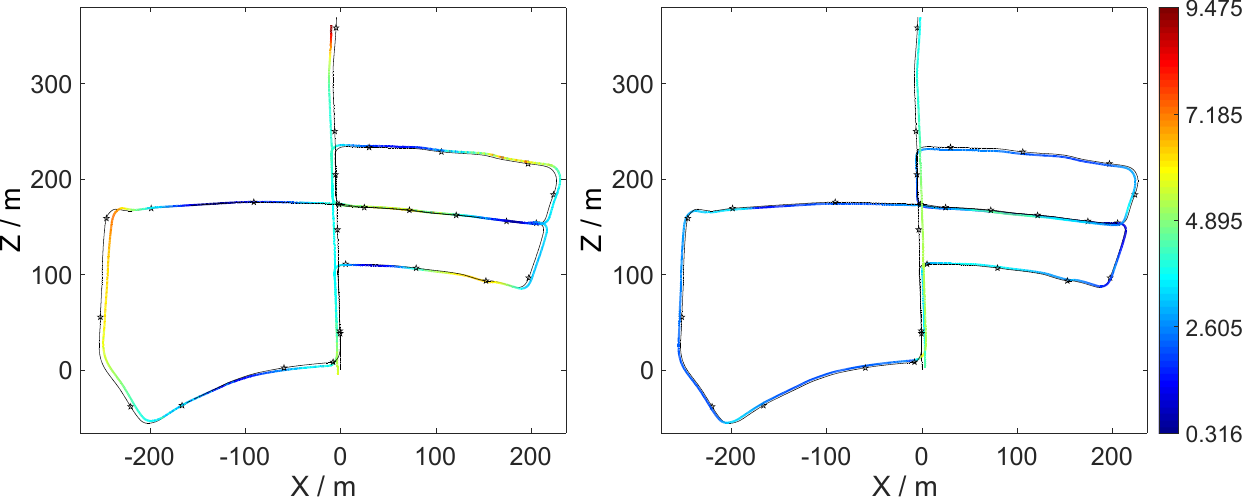}
            }
            \vspace{-10pt}
		\end{center}
		%\caption{Estimated visual odometry trajectories. The left column reports the results of \texttt{ORB-SLAM2}. The right column reports the results of Our \texttt{1AC-SLAM}. Colorful lines are estimated trajectories, and black lines with stars are ground truth trajectories. Best viewed in color.}
		\label{fig:trajectory}
	\end{figure}
	
	\begin{figure}[htbp]
		\begin{center}
			\subfigure[Seq.06]
			{
				\includegraphics[width=1.0\linewidth]{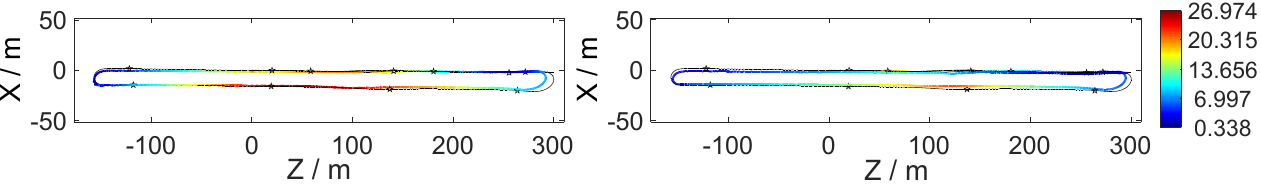}
			}
			\subfigure[Seq.07]
			{
				\includegraphics[width=1.0\linewidth]{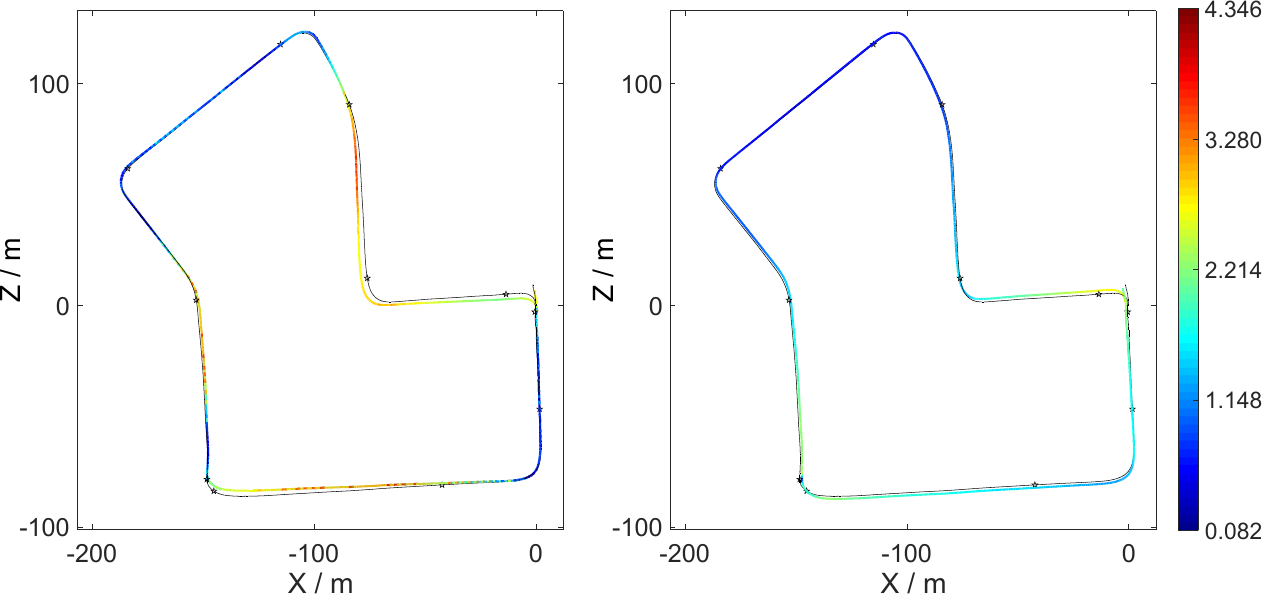}
			}
			\subfigure[Seq.08]
			{
				\includegraphics[width=1.0\linewidth]{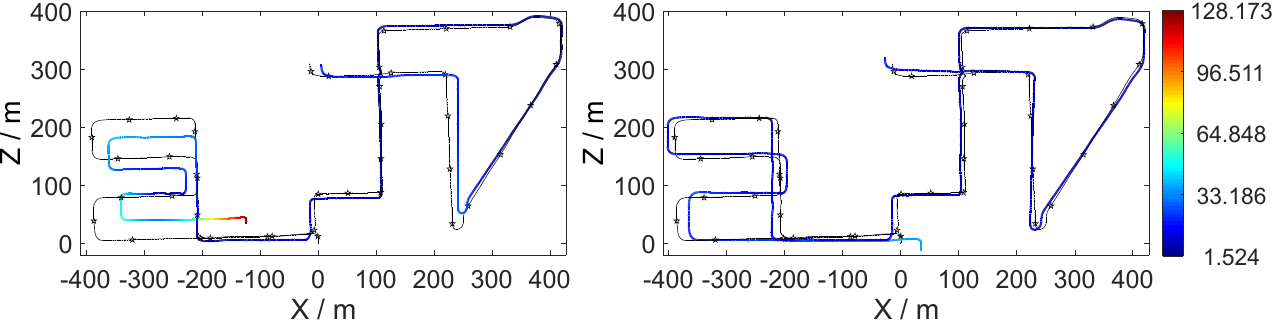}
			}
			\subfigure[Seq.09]
			{
				\includegraphics[width=1.0\linewidth]{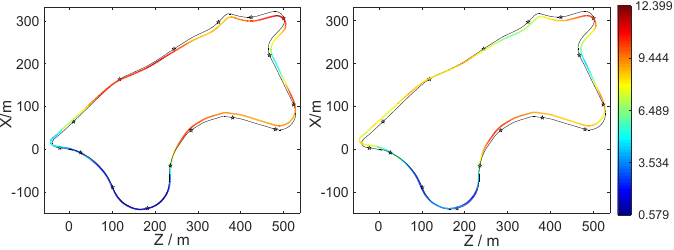}
			}
			\subfigure[Seq.10]
			{
				\includegraphics[width=1.0\linewidth]{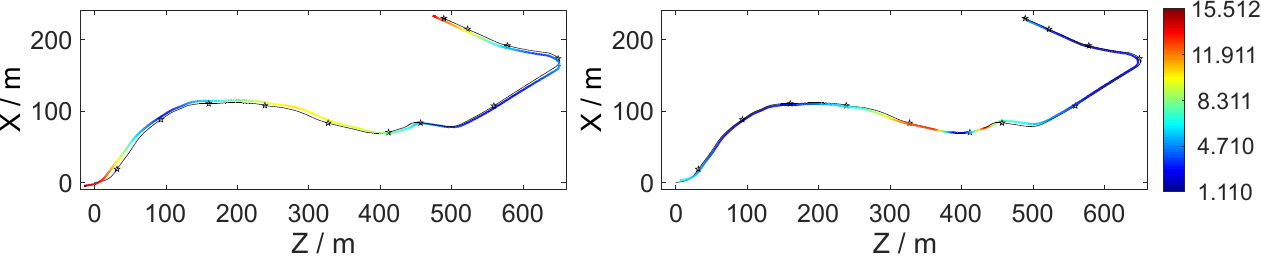}
			}
		\end{center}
		\caption{Estimated visual odometry trajectories. The left column reports the results of \texttt{ORB-SLAM2}. The right column reports the results of our monocular visual odometry. Colorful curves are estimated trajectories, and black curves with stars are ground truth trajectories. Best viewed in color.}
		\label{fig:trajectory}
	\end{figure}

	%\newpage
	{\small
		\bibliographystyle{ieee_fullname}
		\bibliography{myBibGuan}
	}	
\end{document}